\documentclass[acmsmall,screen]{acmart}

\usepackage{times}
\usepackage{soul}
\usepackage{natbib}
\usepackage{float}

\usepackage{url}
\usepackage{graphicx}
\usepackage{physics}
\usepackage{amsmath}
\usepackage{multirow}
\usepackage{amsthm}
\usepackage{booktabs}
\usepackage{algorithm}
\usepackage{algorithmic}
\usepackage{xcolor,color}
\usepackage{hyperref}
\usepackage{tablefootnote}
\usepackage{mathrsfs}

\usepackage{makecell}

\usepackage{framed}
\usepackage{color}
\definecolor{shadecolor}{rgb}{0.92,0.92,0.92}

\usepackage{ulem}

\AtBeginDocument{%
  \providecommand\BibTeX{{%
    \normalfont B\kern-0.5em{\scshape i\kern-0.25em b}\kern-0.8em\TeX}}}

\setcopyright{acmcopyright}
\copyrightyear{2022}
\acmYear{2022}
\acmDOI{XXXXXXX.XXXXXXX}

\acmJournal{CSUR}
\acmVolume{0}
\acmNumber{0}
\acmArticle{0}
\acmMonth{0}

\theoremstyle{plain}
\newtheorem{theorem}{Theorem}

\theoremstyle{definition}

\newtheorem{concept}{Concept}

\theoremstyle{remark}

\begin{document}

\title{A Survey of Quantum-Cognitively Inspired Sentiment Analysis Models}

\author{Yaochen Liu}
\affiliation{%
  \institution{Beijing Institute of Technology}
  \city{Beijing}
  \country{China}
}
\email{yaochen@bit.edu.cn}

\author{Qiuchi Li}
\authornote{Corresponding author}
\affiliation{%
  \institution{University of Copenhagen}
  \city{Copenhagen}
  \country{Denmark}}
\email{qiuchi.li@di.ku.dk}

\author{Benyou Wang}
\authornote{Corresponding author}
\affiliation{%
  \institution{The Chinese University of Hong Kong, Shenzhen}
  \city{Shenzhen}
  \country{China}
}
\email{wang@dei.unipd.it}

\author{Yazhou Zhang}
\affiliation{%
  \institution{Zhengzhou University of Light Industry}
  \city{Zhengzhou}
  \country{China}
}
\email{yzzhang@zzuli.edu.cn}

\author{Dawei Song}
\authornote{Corresponding author}
\affiliation{%
  \institution{Beijing Institute of Technology, Beijing, China \& The Open University, UK}
  \city{}
 \country{}
}
\email{dwsong@bit.edu.cn}


\begin{abstract}
Quantum theory, originally proposed as a physical theory to describe the motions of microscopic particles, has been applied to various non-physics domains involving human cognition and decision-making that are inherently uncertain and exhibit certain non-classical, quantum-like characteristics. Sentiment analysis is a typical example of such domains. In the last few years, by leveraging the modeling power of quantum probability (a non-classical probability stemming from quantum mechanics methodology) and deep neural networks, a range of novel quantum-cognitively inspired models for sentiment analysis have emerged and performed well. This survey presents a timely overview of the latest developments in this fascinating cross-disciplinary area. 
We first provide a background of quantum probability and quantum cognition at a theoretical level, analyzing their advantages over classical theories in modeling the cognitive aspects of sentiment analysis. Then, recent quantum-cognitively inspired models are introduced and discussed in detail, focusing on how they approach the key challenges of the sentiment analysis task. Finally, we discuss the limitations of the current research and highlight future research directions.
\end{abstract}



\maketitle

\section{Introduction}

\subsection{Problem}
Cognition and decision-making under uncertainty are main challenges for artificial intelligence (AI). Humans interact with the environment and communicate with each other, involving a combination of cognitive processes, among which understanding the sentiment underlying the communications is crucial yet challenging. 
Sentiment analysis is a research area that involves recognizing human emotional intelligence~\citep{Ma2021,Esteban2017}. In this paper, we use the term ``sentiment analysis'' in a broader sense that covers the recognition of different types of affect, such as sentiment\footnote{In a narrow sense, sentiment is a thought, opinion, or idea based on a feeling about something or a way of thinking about something (see its definition in Cambridge Dictionary \url{https://dictionary.cambridge.org/dictionary/english/sentiment}). Typical types of sentiment include \textit{positive}, \textit{neutral} and \textit{negative}. }, sarcasm\footnote{Sarcasm is a kind of rhetorical strategy that is intended to express criticism or mock emotions by means of hyperbole, figuration, etc.~\cite{Zhang2021cfn}.}, and emotions\footnote{Emotions are mental states brought on by neurophysiological changes, variously associated with thoughts, feelings, behavioral responses, and a degree of pleasure or displeasure (see its definition on Wikipedia \url{https://en.wikipedia.org/wiki/Emotion}). Typical emotion types include happiness, sadness, fear, and surprise.}.

\subsection{Challenges of Sentiment Analysis}
The main challenge of sentiment analysis mainly resides in its inherent uncertainty. For example, the sentence ``{\tt You're safe.}'' states that \textit{the object is not in danger now}. Observing the sentence alone, one cannot tell whether the speaker is happy or relieved since his or her sentiment might also be neutral or positive. When the sentence is preceded by ``{\tt Zombies eat brains.}'', a clear ``negative'' sentiment and ``disgust'' emotion are expressed in this \textit{context}, and the uncertainty is eliminated. 

The second challenge is that humans may express feelings in \textit{multiple modalities}, e.g., a mixture of textual language, facial expression and acoustic tone~\citep{Chen2020}. Different modalities can carry different information that may be complementary or contradictory, and the final judgment on the emotion or sentiment should be made by jointly considering all the modalities. Hence, contextuality and multimodal complementarity construct the inherent uncertainty of human affects, making sentiment analysis a challenging task.

The third challenge is the interdisciplinary nature of sentiment analysis, which is, by definition, a task spanning the disciplines of computing and cognitive science. Most existing work considers sentiment analysis as a pure computing task but neglects its cognitive aspect. We believe that effectively understanding sentiment in a way that is compatible with human cognition is critical to building more human-like AI technologies.

\subsection{Cognitive Aspect of Sentiment Analysis}
 Cognitive science is an interdisciplinary study of mind and intelligence. It studies what cognition is, what it is used for and how it works, and how information is expressed as sensation, language, attention, reasoning and emotion. Sentiment analysis, by nature, involves a range of complex cognitive processes. Typical questions it aims to answer include, e.g., \textit{can we make judgments jointly over sentiment, emotion and sarcasm?}, \textit{How do the judgments over multiple modalities interact with each other to reach a final decision?}, \textit{How do the conversational and modality contexts change our cognitive state to affect the sentiment decision?}

 \textbf{\textit{Cognition and Classical Probability.}} Classical (Kolmogorovian) probability theory is at the core of current cognitive and decision models, such as Bayesian and connectionist networks. The classical probability is based on the principle of unicity ~\cite{Griffiths2003}, where a universe sample space is assumed to provide a complete and exhaustive description of all events (as subsets) that can occur in an experiment. These events are commutative (i.e., the order of events will not affect the outcome). The principle of unicity complies with set theory and Boolean algebra and oversimplifies the complex nature of our cognitive and decision-making processes. For example, the outcomes of a real-life cognitive process may be sampled from several separate sample spaces (e.g., incompatible questions that affect each other so that the order matters). In this case, the commutative axiom is violated, so the joint probability of events does not hold without considering their order.  
 Furthermore, according to Boolean algebra, the distributive axiom and the law of total probability always hold. However, our cognitive process is highly subtle and complex in the sense that all possible states can be ambiguous (e.g., positive, negative or neutral sentiment polarities) simultaneously, which also interfere with each other. The existence of such a ``superposed'' state and interference effect violates the law of total probability ~\cite{Busemeyer2012}.  
Therefore, the classical probability is insufficient to model human cognition. This is further explained in \textit{Section 2} and \textit{Section 3.1}.

\textbf{\textit{Cognition and Artificial Life.}} Artificial Life (ALife) is an interdisciplinary field of research that aims to understand the fundamental principles of living systems by creating and studying artificial lifeforms. This field is concerned with the development of computational models that can simulate the behavior and evolution of living organisms, as well as the design and implementation of physical systems that exhibit life-like characteristics.
ALife draws inspiration from a wide range of fields, including biology, physics, computer science, robotics, and philosophy ~\citep{Bedau2003, Pfeifer2007}.
ALife puts forward a definition of cognition: "cognition emerges from 
the interaction of a sensing system, an artificial brain and a motor system".
 In the definition, a sensing system can involve different kinds of sensors 
in robotic intelligent agents, 
and an artificial brain can be an ML algorithm ~\citep{Pfeifer2001}. This definition plays an important role in the area of artificial intelligence.

Braitenberg Vehicles, important robotic agents in the field of ALife designed by Braitenberg in 1984~\citep{Braitenberg1986}, are commonly used to study animal behavior, particularly in the areas of cognition and perception. They are simple robots driven by two or more sensors, which process perceived information to produce a series of action responses. Braitenberg Vehicles and ALife are very helpful for improving the interpretability of ML approaches.

\subsection{Classical Solutions}
 Most current sentiment analysis models are focused on addressing the first two challenges based on classical machine learning and deep learning techniques.

Machine learning (ML) automatically analyzes and obtains patterns from data and uses them to predict unseen data ~\cite{Jordan2015}. It has been successfully applied to a wide range of tasks, such as classification, clustering, and representation learning. In particular, taking inspiration from the biological brain with neurons and their connections, artificial neural networks (ANNs) were designed ~\cite{Liu2017}. Furthermore, based on multiple ANN layers, deep learning (DL) ~\cite{Dong2021} has achieved state-of-the-art (SOTA) performance in many applications (including sentiment analysis) due to its powerful learning capability.
 
 Many attempts (e.g., the encoder-decoder architecture) have been made to ``translate'' the representation of a single sentence under one modality to the representation of another modality and take the hidden unit as the joint representation of the sentence~\cite{Pham2019, Tsai2019}. To suit the multimodal context, approaches based on pretrained language models have recently been proposed to integrate visual and acoustic features into pretrained word-level textual features ~\cite{Rahman2020}. In addition to multimodal interaction, contextual information also plays an important role. Recurrent neural networks (RNNs) and their variants (e.g., GRU and LSTM) have been widely used to deal with sequential information in conversations~\cite{Ghosh2017, Poria2017}. The attention mechanism has also been applied to study contextual interaction~\cite{Kumar2020}.

It is important to stress that classical statistical mechanics has had a great influence on ML. 
"Statistical mechanics deals with large systems of stochastically interacting microscopic elements (particles, atomic magnets, polymers, etc.)" ~\citep{Coolen2001}. It studies a system at a macroscopic rather than at a microscopic level. In other words, statistical ensembles are assigned to N particles in classical statistical 
mechanics.
Many ML approaches are being theoretically analyzed using methods borrowed from mathematics and theoretical physics, especially statistical mechanics. This can result in the development of improved algorithms or a deeper understanding of the conditions necessary for achieving excellent performance ~\citep{Gabrie2019}. 
For example, 
Coolen discussed the statistical mechanical analysis of neuronal firing processes in recurrent neural  networks with static synapses from a network operation, rather than a network learning, perspective ~\citep{Coolen2001, Coolen2001a}.
In addition, ML algorithms have been inspired and are still being influenced by methods that are derived from statistical mechanics, e.g., the use of mean-field theory and its variations. Over the past few years, ML methods have also gained attention for their effectiveness in addressing problems in statistical and theoretical physics, for example, the automated identification of matter phases, the development of efficient representations of quantum wave functions, and the study of non-linear mechanical behaviors of dissipative materials~\citep{Agliari2020, Danoun2022}.
On the other hand, various key concepts in ML
 draw heavily on ideas and methods from classical statistical mechanics. For example, neural network models, Gaussian process models and decision tree models can be described and optimized in terms of probabilistic statistics.
Optimization algorithms in machine learning, such as gradient descent, conjugate gradient 
descent,
 and Newton-like methods, can also be seen as extensions and applications of some optimization methods from classical statistical mechanics. For example, the gradient descent algorithm can be seen as a special case of the simulated annealing algorithm.
More recently emerging ML techniques, such as deep learning and reinforcement learning, have also been inspired and influenced by classical statistical mechanics. For example, convolutional neural networks can be seen as a generalization of multilayer perceptrons, while the Markov decision processes in reinforcement learning can be seen as the modeling and optimization of stochastic dynamical systems ~\citep{Mehta2018, Weng2022}.

In addition to classical statistical mechanics, ALife has also helped improve the interpretability of ML methods. ~\cite{Iizuka2004} described a study about turn-taking behavior-based human-designed agents. In this work, agents were designed with recurrent neural networks (RNNs) to perform turn-taking behavior. From the ALife perspective, RNNs are similar to "brains". The constituent elements of the RNNs are given their own meanings, offering the proposed approach a solid ontological basis. These agents can be regarded as RNN-enhanced Braitenberg vehicles: they use complex rules (an ML algorithm) to generate decisions based on the behavior of other agents and their own contexts.

Even when theoretically equipped with statistical mechanics or ALife, the classical machine learning models fail to address the cognitive aspects of sentiment analysis.
As shown in the previous subsection, the classical probability theory underlying these classical solutions is insufficient to model complex cognitive processes. With the help of statistical mechanics and ALife, the interpretability of classical solutions can be enhanced to a certain extent. Nevertheless, a non-classical cognitive perspective with a systematic set of explanations is needed to achieve a better theoretical grounding for more effective sentiment analysis models.

\subsection{Quantum-Cognitively Inspired Solutions}

Recent research in cognitive science and experimental psychology has revealed that human cognition and decision making, to a certain extent, can exhibit non-classical, quantum-like behaviors that can be better explained with quantum theory (QT) and the non-classical probability of quantum mechanics methodology (PQMM) ~\citep{Bruza2015, Busemeyer2012}. These findings provide the opportunity to use quantum theories to solve problems related to cognitive science. 

In this work, we use the expression ``quantum probability'' (QP) as equivalent to PQMM. 

\subsubsection{From Quantum Mechanics to Sentiment Analysis}
The relationship between quantum mechanics and sentiment analysis is not straightforward. First, we should consider theoretical mechanics and statistical mechanics, which inspired the emergence of quantum mechanics. Then, we need to make clear the road map from quantum mechanics to sentiment analysis.

The first step from quantum mechanics to sentiment analysis is related to the game theory, especially the thermodynamic description of game theory. 
The introduction of quantum mechanics into classical game theory gave rise to quantum game theory, which applies the principles and methods of quantum mechanics to the analysis and modeling of game behaviors~\cite{Brandenburger2010}.
Thermodynamics is closely related to quantum mechanics and has been used to answer fundamental questions in the discipline of quantum thermodynamics~\cite{Ali2020}.
Probability distribution functions and statistical methods are the basis of the thermodynamic description of game theory, and the wave function in quantum mechanics can also be regarded as a probability distribution function. Therefore, quantum mechanics can be used to study the thermodynamic description of game theory ~\citep{Pomorski2023}.

Second, game theory and the thermodynamic description of game theory have an impact on ALife. Cellular automata, tools in ALife that are used to simulate the behavior and evolution of life, can be described by the concepts of thermodynamics \cite{Pomorski2023}. For example, in simulations, we can use thermodynamic variables, such as entropy and temperature, to describe the evolution and stability of a system. 
In addition, game theory can be applied to the study of artificial life. Through game theory, we can study the interactions and competition between different individuals in a living system and how they gain advantage through game strategies. These studies can aid understanding of the  evolution of living systems and also provide references for the design and development of artificial life ~\citep{Wolfram1983,Badii1997}. 

Fundamentally speaking, quantum-cognitively inspired sentiment analysis models are implemented based on machine learning frameworks (e.g., Tensorflow and PyTorch). The last step is determining the relationship between the ALife and ML approaches, which was introduced in the last subsection (the relationship between quantum mechanics and machine learning will be discussed later). 

\subsubsection{Underlying Mathematical Grounding: Quantum Probability}

One of the most interesting features of QT lies in how probability is computed. Different from the classical probability, QP operates on subspaces are not necessarily commutative events. It builds on the basis of the von Neumann axioms ~\cite{VonNeumann2018} and does not always obey the law of total probability 
\footnote{
The theoretical support for this comes from dissipative quantum systems. Dissipative quantum systems are quantum systems that gradually lose coherence and tend towards thermal equilibrium due to irreversible processes, such as the loss of energy and information, through coupling with the environment in an open system. A characteristic of quantum dissipative systems is that the normalization of the quantum state is not preserved due to the coupling with the environment, which leads to the continuous loss of information and energy from the system. As a result, the evolution of the quantum state is no longer unitary; in other words, the total probability is not conserved.
}
, which is an important axiom in classical probability and the foundation for inferences with Bayesian networks ~\cite{Khrennikov2010}.
Therefore, with QP, incompatible questions cannot be evaluated on the same basis, so they require separate sample spaces. QP allows partial Boolean algebra\footnote{One sample space can be used to answer a first set of questions, and another sample space can be used to answer a different set of questions using a Boolean approach, but both Boolean subalgebras are pasted together in a coherent but non-Boolean way.}, and can better explain the nonclassical cognitive phenomenon that violates the law of total probability. For instance, the law of total probability is violated in the disjunction experiment and the decision-making experiment in psychology, as in the double-slit type of experiments in quantum physics~\cite{Busemeyer2012}. Thus, QP has been used to describe and explain puzzling empirical findings in human cognitive processes and emotional activities ~\cite{Busemeyer2015}. In this paper, nonclassical cognitive phenomena that violate the axioms of classical probability theory but can be better explained by QP are referred to as quantum-like cognitive behaviors. A detailed explanation is given in \textit{Section 2}. 

\subsubsection{Motivation: A Quantum Cognition Perspective}
QT has been applied to various non-physical domains,
such as game theory, management science, and finance and economics. ~\citep{Drabik2011, Pomorski2023}. Recently, it has been used in research areas
involving human cognition and decision-making, resulting in a new research community loosely recognized as \textbf{\textit{quantum cognition}}. Currently, there is no precise scientific definition of quantum cognition. Generally, it posits that human cognitive information processing takes place in a quantum-theoretical way, supported by increasing literature from cognitive science and empirical psychology findings. Accordingly, it advocates the use of mathematical principles from QT and QP as a new conceptual framework and a coherent set of formal tools to help formalize and understand cognitive systems and processes and to better explain and model nonclassical, quantum-like cognitive behaviors \cite{Busemeyer2015, Bruza2015, Busemeyer2014}. See \textit{Section 3.1} for a more detailed description.

\textbf{\textit{Examples of quantum-like cognitive behaviors in sentiment analysis.}}
The multimodal sentiment analysis task has no clear boundaries among different types of human affects, e.g., sentiment, emotion and sarcasm. Indeed, they are often correlated, and their decisions cannot be made simultaneously in disregard to their relative order. This is analogous to the incompatible measurements in QT. Moreover, sentiment judgments over multiple modalities (such as textual, acoustic and visual) interfere with each other in a complex way, and such interference affects our final sentiment decision. This is similar to the quantum interference effect in which two particles pass through two slits, interfering with each other and affecting their final position on the detection screen. See \textit{Section 3.2} for more details.

 Accordingly, \textbf{\textit{a quantum-cognitive perspective}} for sentiment analysis has recently been proposed, where QT formalism is used to model a range of key cognitive issues (e.g., the user's incompatible cognitive states underlying judgments on different sentiment types, the nonseparability (entanglement) of multimodal information, and the interference effect in the fusion of decisions over multiple modalities) ~\citep{Gkoumas2021, Gkoumas2021a, Zhang2020}. See \textit{Section 3.2} for a more detailed description. 
 
 Based on this perspective, a range of quantum-cognitively inspired models have been developed, which are surveyed in this paper. Practically, these models are implemented by integrating quantum modeling of certain cognitive issues with deep ML architectures.

\subsubsection{QT and ML}

In the last two decades, the field of quantum computing (QC) has been significantly developed. QC uses QT concepts and mechanisms, such as quantum superposition, entanglement and measurement, to accelerate computation and enhance storage capacity~\cite{MartinGuerrero2022, Houssein2022}. In addition, ML algorithms have also been designed based on QT, called quantum machine learning (QML) ~\cite{Biamonte2017, Schuld2015}. QML algorithms can be classified into two paradigms ~\cite{Houssein2022}. One is the QC paradigm, which usually implements algorithms on QC devices or implements quantum circuits on classical computers to run the algorithms ~\cite{Rebentrost2014}. The most significant advantage of this paradigm is its speedup over classical ML. The second paradigm is quantum-inspired ML (QiML) ~\cite{Rebentrost2014}, which is based on the mathematical formalism of QT and implemented on classical ML and DL frameworks. 
In this paper, we are concerned with the second paradigm, which aligns with the quantum-cognitive perspective of sentiment analysis mentioned above.

\subsubsection{Implementation: Quantum-Cognitively Inspired Models and Artificial Neural Networks}

Different from QC or QML, the to-be-surveyed quantum-cognitively inspired sentiment analysis models are intended for classical sentiment analysis tasks using QP and operate on classical computers. Almost all implementations of the surveyed models are based on ANN architectures. 
In fact, ANN architectures have been widely used in physics. For example, ~\cite{Danoun2022} explored the capability of a hybrid physics--RNN model (a Thermodynamically Consistent Recurrent Neural Network, or ThC-RNN) as a dependable and consistent replacement for the constitutive modeling of dissipative materials.
Implementing quantum-inspired components and designing training frameworks based on ANNs is a widely used paradigm in this research area. The motivations are threefold:

First, classical neural networks have a certain degree of quantum information processing capability ~\cite{Vanchurin2020, Jonsson2018}. This is not limited to using ANNs to implement the quantum inverter operator and CNOT operation or using ANNs to simulate QC ~\cite{Jeswal2019, Ezhov2000, Faber2002}. We can also find some quantum-like characteristics, for example, entanglement, in ANNs ~\cite{Deng2017, Levine2018}. Indeed, there have been attempts to establish connections between quantum mechanics and neural networks 
For example, 
by using Madelung equations and Hamilton--Jacobi equations, ~\citep{Vanchurin2020} showed that neural networks are capable of displaying both quantum and classical behaviors. 
~\citep{Landman2022} introduced two quantum neural network approaches to medical image classification tasks and achieved considerable effectiveness while demonstrating better scalability. 
~\citep{Carrasquilla2021} reviewed three applications of ANNs in condensed matter physics and quantum information.
Traditional quantum many-body problems are computationally  difficult, so researchers have tried to use neural networks to approximate these problems. The reason why neural networks can reproduce quantum mechanical experiments is that neural networks incorporate and can represent many classical statistical mechanics systems that may be equivalent to QM to a large extent.
Table~\ref{table-quantumconcept-nn}~\cite{Ezhov2000} lists the main concepts of quantum mechanics and neural networks. Establishing correspondences between two fields (quantum mechanics and neural networks) is a major challenge. ~\cite{Perus1996} noted that the decoherence of a quantum state can be considered an analog of the neural network state evolution to an attractor basin. ~\cite{Behrman1996} developed a temporal model of a quantum neural network in which the temporal evolution of a system resembles the equations for virtual neurons. ~\cite{Chrisley1995} regarded the positions of slits in the interference experiment as neuron state values, and the positions of other slits encode the network weights. Inspired by Everett's many universes interpretation of quantum mechanics ~\cite{EverettIII1957}, ~\cite{Menneer1995} designed a set of multilayered perceptrons. Each layer is trained on only one pattern, and all layers are combined into a quantum network in which weights are superpositions of the weights of all perceptrons existing in the parallel universes. A summary of the aforementioned works is listed in Table~\ref{table:quantum-analogies-nn}~\cite{Ezhov2000}. 

Second, from a theoretical point of view, the quantum-cognitively inspired models and ANNs share a common cognitive perspective, where sentiment analysis, the surveyed models task, also manifests. The ANNs, to some extent, are also cognition-inspired, in the sense that they were designed to mimic human brain functioning in perception and cognition~\cite{Liu2017}. Human decision-making and cognition might arguably follow a quantum or quantum-like fashion ~\cite{Fell2019, Bruza2015, Busemeyer2012}. Furthermore, QT (and QP) is more general than classical and statistical physics (and classical probability). Thus, it is natural to adopt QT as a general theory to describe the world.
The to-be-surveyed models and the works on designing quantum circuits with trainable parameters are natural methods to bridge QT and ANNs.

Third, from a methodological point of view, the connections between QT and ANNs have been investigated in both directions. While ANNs have been used for parameterizing ~\cite{Carleo2017} and estimating ~\cite{Torlai2018, Schmale2022} quantum states, deep neural networks provide an effective parameter learning and model training mechanism, which allows end-to-end training of quantum-inspired models.

In summary, the connections between QT and ANNs on different levels motivate research into quantum-inspired neural networks for sentiment analysis. We follow the research in cognitive science and experimental psychology, which states that quantum-like phenomena (e.g., interference, entanglement, order effect, and incompatibility) in human cognition and decision-making processes that can only be explained by QP theory. Experimental results from surveyed papers demonstrate that QT formalism, in combination with deep neural networks, can effectively handle the multimodal sentiment analysis task.

\begin{table*}[th]
\begin{center}
\caption{\label{table-quantumconcept-nn} Main concepts of quantum mechanics and neural networks. This table is reproduced from Table 1 in Chapter 11 of ~\cite{Ezhov2000}. }
\scalebox{1}{
\begin{tabular}{c|c}
\hline
\textbf{Quantum Mechanics}        & \textbf{Neural Networks}               \\ \hline
Wave function            & neuron                        \\ \hline
Superposition (coherence) & interconnection (weights)      \\ \hline
Measurement (decoherence) & evolution to attractor        \\ \hline
Entanglement             & learning rule                 \\ \hline
Unitary transformations  & gain function (transformation) \\ \hline
\end{tabular}}
\end{center}
\end{table*}

\begin{table*}[th]
\begin{center}
	\caption{ Quantum analogies used for different concepts of artificial neural networks. This table is reproduced from Table 2 in Chapter 11 of ~\cite{Ezhov2000}.}
	\label{table:quantum-analogies-nn}
    \scalebox{0.85}{
    \resizebox{\linewidth}{!}{
    \begin{tabular}{c|c|c|c|c|c}
\hline
\textbf{Model}                 & \textbf{Neuron}                   & \textbf{Connections}                  & \textbf{Transformation}                                           & \textbf{Network}                                & \textbf{Dynamics}                                \\ \hline
\makecell[c]{Perus ~\cite{Perus1996}}                 & \makecell[c]{quantum}                  & \makecell[c]{Green \\ function}               & \makecell[c]{linear}                                                   & \makecell[c]{temporal}                               & \makecell[c]{collapse as \\convergence\\ to attractor}    \\ \hline
\makecell[c]{Chirsley ~\cite{Chrisley1995}}              & \makecell[c]{classical \\(slit \\ position)} & \makecell[c]{classical\\(slit \\ position)}     & \makecell[c]{nonlinear \\through \\ superposition}                          & \makecell[c]{multi-layer}                             & \makecell[c]{non- \\superposit-\\ional}                     \\ \hline
\makecell[c]{Behrman ~\cite{Behrman1996}}        & \makecell[c]{time slice,\\ quantum}      & \makecell[c]{interactions \\through\\ phonons} & \makecell[c]{nonlinear\\ through\\ potential energy \\and exponent\\ function} & \makecell[c]{temporal\\ and\\ spatial}                   & \makecell[c]{Feynman \\path integral}                   \\ \hline
\makecell[c]{Menneer \\and\\ Narayanaa ~\cite{Menneer1995}} & classical                & classical                    & nonlinear                                                & \makecell[c]{single-item \\networks\\ in many\\ universes} & classical                               \\ \hline
\end{tabular}
    }}
\end{center}
\end{table*}

\subsubsection{Overview of the Models to be Surveyed}
In the last few years, leveraging the mathematical formalism of QT and the learning capability of deep neural networks, a range of novel quantum-cognitively inspired sentiment analysis models have emerged and achieved significant performance. 

QT has been used to build more comprehensive representations. 
An early model, quantum-inspired sentiment representation (QSR) ~\cite{Zhang2019}, extracted sentiment phrases, adjectives and adverbs as projectors. Sentences were then encapsulated in density matrices constructed by the projectors with the maximum likelihood estimation prior for sentiment classification. 
Another model, the quantum-inspired model based on a convolutional neural network (QI-CNN) for sentiment analysis ~\cite{Li2021a}, leveraged the concept of a density matrix to encode the mixture of semantic subspaces and capture the feature interactions between sentiment words. Then, a convolutional neural network (CNN) was used to obtain the sentiment classification result. 

In addition, QT has shown an innate advantage in modeling conversational and multimodal interactions.
~\cite{Zhang2019a} proposed a model named quantum-inspired interactive networks (QIN), which leveraged both the mathematical formalism of QT and the long short-term memory (LSTM) structure to jointly capture intrautterance and interutterance interactions. 
~\cite{Guo2019} designed a novel attention mechanism called DMATT, which is combined with a bidirectional gated recurrent unit (Bi-GRU), resulting in a model called quantum-inspired DMATT-BiGRU. The static density matrix and dynamic density matrix were introduced to capture the intrautterance and interutterance interactions, which are input to the GRU and the attention layer, respectively.
~\cite{Zhang2021cfn} designed a complex-valued fuzzy network (CFN) for sarcasm detection in conversations. In CFN, a quantum composite system is constructed by describing the contextual interactions between adjacent utterances as the interactions between a quantum system and its environments.
~\cite{Zhang2018} proposed a quantum-inspired multimodal sentiment analysis (QMSA) framework, which is composed of a quantum-inspired multimodal representation (QMR) model and a multimodal decision fusion strategy inspired by quantum interference (QIMF).
~\cite{Gkoumas2021} introduced a novel decision-level fusion strategy (in abbreviation DecisionFusion) inspired by quantum cognition, which learns incompatible cognitive states underlying sentiment decision-making under different modalities.
Inspired by the quantum entanglement phenomenon, ~\cite{Gkoumas2021a} proposed a quantum probabilistic neural model called the entanglement-driven fusion neural network (EFNN). 
~\cite{Li2021} developed a quantum-inspired multimodal fusion (QMF) model that formulated the word interactions within one modality as quantum superposition and the interaction across modalities as quantum entanglement at different stages.
As an extension of the QIN model ~\cite{Zhang2019a}, ~\cite{Zhang2020} proposed a quantum-like multimodal network (QMN) framework by leveraging various QP formalisms and the LSTM architecture.
~\cite{Li2021b} designed a quantum-inspired neural network (QMNN) for multimodal conversational emotion recognition. 
As an extension of the CFN model~\cite{Zhang2021cfn}, ~\cite{Liu2021} proposed a quantum probability-based multitask learning framework (QPM) for joint multimodal sentiment and sarcasm classification, in which the interaction among modalities is modeled as quantum interference while the sentiment and sarcasm decisions are modeled as incompatible measurements. 

These quantum-cognitively inspired models have achieved promising results compared with the SOTA traditional models while introducing a better capability to explicitly model the quantum-cognitive effects and offering an improved interpretability. These models are comprehensive reviewed in \textit{Section 4}.

\subsubsection{Advantages of Quantum-Cognitively Inspired Models}

We conclude this section with the benefits of using quantum-cognitively nspired approaches: 1) QT provides a new method to unify information representation, decision-making, and interaction in a single probabilistic space. Here, we make use of the QT mathematical formalism to probabilistically describe the world (e.g., using events, probability and decision-making). 2) There exist some nonclassical, quantum-like behaviors in sentiment analysis, which can be better explained by QT and QP. Therefore, these quantum-cognitively inspired models are expected to lead to better sentiment analysis performance. 3) The experimental results show that in some scenarios, these models can improve the performance while providing better interpretability ~\cite{Zhang2021cfn, Li2021, Zhang2019a}. 4) Although these quantum-inspired models are currently implemented on classical deep neural network structures and run on classical computers, they can be extended to QC hardware in the future ~\cite{Meichanetzidis2020}. Therefore, the surveyed models lay a foundation for future exploration of the interplay between QC and quantum-inspired models. 

\subsection{Scope and Significance of This Survey}

Quantum-inspired models have been explored for nearly two decades. Early efforts mainly focused on information retrieval following the initial inspiration from Keith van Rijsbergen's seminal book ~\citep{Rijsbergen2004}. A wide range of models have been developed for information retrieval~\citep{Sordoni2013,Wang2018, Jiang2020, Zhang2018a}, question answering~\citep{Guo2021, Zhang2018b}, document representation~\citep{Wang2019}, and image representation~\citep{Dang2018}. There are some existing surveys~\citep{Wu2021,Yan2016,Uprety2020,Abohashima2020} for these research areas, but none focused on the quantum-cognitive modeling of sentiment analysis. Indeed, the main breakthroughs in this area were made in the last three to four years and need to be comprehensively and critically reviewed to clarify the current landscape of this emerging area and draw a roadmap for further development. 

We aim to fill this gap by surveying quantum-cognitively inspired models for sentiment analysis. We hope this survey will benefit not only the further study of quantum-inspired models but also other relevant natural language processing (NLP) domains, e.g., multimodal language analysis.  

We need to emphasize that this survey focuses on practical sentiment analysis models that are quantum-cognitively inspired instead of cognitive science and cognitive models. We introduce the concept of quantum cognition for readers to better understand the motivation of the surveyed models. These quantum-inspired models do not assume that the human brain works quantumly. Instead, they are motivated by the findings that QP provides a better method for explaining cognitive processes.

We would also like to clarify that QC and QML that simulate quantum circuits or run on QC devices are beyond the scope of this paper. Instead, the surveyed models are built on top of the mathematical formalism of QT and implemented on classical ANNs architectures. They are intended for classical problems using QP and operate on classical computers.

\subsection{Structure of This Survey}
The rest of the paper is organized as follows. Section 2 introduces the preliminaries of QT and QP. Section 3 overviews research in quantum cognition and illustrates the advantages of taking a quantum-cognitive perspective and applying QT in sentiment analysis tasks. Section 4 provides an in-depth review of the recent advances in quantum-cognitively inspired sentiment analysis models. Finally, we summarize the reviewed models in Section 5 and provide future research directions in Section 6.

\section{Quantum Theory Preliminaries}
The QT formalism offers a mathematical and conceptual framework for describing microscopic particle behaviors that are intrinsically uncertain. In this section, we introduce the key concepts of QT utilized in quantum cognition that are related to quantum states (in Section~\ref{sec:states}), measurements (in Section~\ref{sec:measurement}), interference and entanglement (in Section~\ref{sec:interence_entanglement}), and probabilistic interpretation (in Section~\ref{sec:probability}).

\subsection{Quantum States}
\label{sec:states}

The mathematical formalism of QT is established on a complex Hilbert space, denoted as $\mathcal{H}$. The simplest quantum system is a qubit, with two possible states represented by the Dirac notation: $\ket{0}$ and $\ket{1}$. A quantum state vector $u$ is expressed as a \textit{ket} $\ket{u}$, and its conjugate transpose is a \textit{bra} $\bra{u}$. The inner product and outer product of two state vectors $\ket{u}$ and $\ket{v}$ are denoted as $\bra{u}\ket{v}$ and $|u\rangle\langle v|$, respectively.

Before introducing the key concepts of QT and how they are used in quantum-cognitively inspired models, we first present several basic QT principles ~\cite{Dirac1981}:

\begin{enumerate}
        \item Various parallel states are simultaneously encoded in a state vector $\ket{\psi}$ that is normalized. See the concept of quantum superposition. 
         
    \item Every observable $A$ is associated with a self-adjoint (Hermitian) operator $\hat{A}$. The possible outcomes for a measurement of observable A can only be the eigenvalues of $\hat{A}$, and the probability of an outcome is given by Born's rule (see the next subsection).
   
    \item Partial or full collapse of the quantum system is caused by weak or strong measurement. This has been used in a quantum-inspired model for conversational sentiment analysis that captures the strong/weak influence between speakers through strong/weak measurement ~\cite{Zhang2019a} (detailed in Section 4.2).

             \item System evolution is in accordance with equations of motion $i\hbar\frac{d}{dt}\ket{\psi} = \mathcal{H}\ket{\psi}$. This was used in a quantum-inspired model for multimodal sentiment analysis that incorporates nonseparable (entangled) cross-modal interactions through quantum evolution ~\cite{Gkoumas2021a} (detailed in Section 4.3).   
    
\end{enumerate}
\begin{concept}
(\textbf{Quantum Superposition}).
A \textit{general} state \footnote{In quantum physics, a state is a mathematical representation where a probability distribution that provides the outcome of each possible measurement on a system is encapsulated. } of a physical system is the combination of all its configurations (a.k.a. superposition), even though the physical system can be in any one of these configurations (e.g., arrangements of particles or fields). The probability in each configuration is determined by a complex-valued weight.
\footnote{https://en.wikipedia.org/wiki/Quantum\_superposition}
\end{concept}

\textit{Quantum superposition} states that a pure quantum state can be in multiple mutually exclusive basis states, e.g., $\ket{0}$ and $\ket{1}$, simultaneously, denoted as $\ket{u} = \alpha \ket{0} + \beta \ket{1}$, with probabilities satisfying $\alpha^2 + \beta^2 =1$. 

\begin{concept}
(\textbf{Pure and Mixed States}).
 A state that cannot be written as a mixture of other states \footnote{Quantum states are composable in the sense that a mixture of quantum states is also a quantum state.} is called a \textit{pure state}, while all other states are called \textit{mixed states}. \footnote{https://en.wikipedia.org/wiki/Quantum\_state}
\end{concept}

A pure quantum state can be represented by a ray in Hilbert space over complex numbers \cite{Bransden1989, Weinberg1995}, while a mixed state cannot.
Drawing an analogy to the sentiment polarity of a word (as a \textit{pure state}), it can be both positive and negative simultaneously until a particular listener judges it under a certain context.
A \textit{quantum mixture} of pure states gives rise to a mixed state represented by a density matrix, $\rho=\sum_i p_i |u_i\rangle\langle u_i|$, where $p_i$ denotes the probability distribution of pure states. Quantum mixture is often used to represent textual documents from different granularities; e.g., a sentence can be represented as the mixture of words it contains as the corresponding pure states.

\subsection{Measurement}
\label{sec:measurement}

\begin{concept}
(\textbf{Quantum Measurement}).
In quantum physics, a measurement is used to quantitatively obtain the probability for a given outcome by manipulating a physical system. The predictions are, by definition, probabilistic.
\footnote{https://en.wikipedia.org/wiki/Measurement\_in\_quantum\_mechanics}
\end{concept}

\textit{Quantum Measurement} is described by a set of measurement operators, denoted as $\left \{ M_{m} \right \}$ acting on the state space of the system being measured, where $m$ represents the possible measurement outcomes. The measurement operators satisfy $\sum_m {M_{m}^{\dagger}M_{m}}= I$, where $\dagger$ denotes the conjugate transpose of a complex matrix. Each operator and the measurement outcome together form an \textit{observable} that determines the measurement. Assume the quantum system is in a state of $\ket{u}$ prior to a measurement; then, the probability of obtaining the outcome $m$ after the measurement is $p\left ( m \right )= \langle u|M_{m}^{\dagger }M_{m}|u\rangle$, as given by Born's rule~\citep{Born1926}. Due to the mathematical constraint on the operators, $\{p(m)\}$ are nonnegative real values that sum to 1. Different from the classical case, quantum measurement causes the system state to \textit{collapse} onto the corresponding state described by the operator $M_{m}$. 
A commonly used type of measurement is \textit{projective measurement}, where the measurement operators $\left \{ M_{m} \right \} = \ket{m}\bra{m}$ are mutually independent and have rank one. This simplifies the calculation of probability to $p\left ( m \right )= |\braket{u}{m}|^2$. After the measurement, the system state is changed to $\ket{m}\bra{m}$ or equivalently $\ket{m}$ at probability $p\left ( m \right )$.

\begin{concept}
\textbf{(Quantum Incompatibility)}.
 Quantum incompatibility indicates that the measurements cannot be accessed jointly without interfering with each other. The order of the measurements must be specified before obtaining the outcomes. 
 \end{concept}

Quantum incompatibility describes the property of a set of quantum measurements. Its roots go back to Heisenberg's uncertainty principle and Bohr's notion of complementarity \cite{Heinosaari2016}. 
A set of observables is \textit{compatible} if measuring with one observable has no influence on the measurement result with another. In this case, it is not necessary to specify the order in which the observables are measured -- they always lead to identical outcomes. Therefore, the observables are also called \textit{commuting observables} when they are compatible. Mathematically, a set of observables are compatible if and only if they share a common set of eigenvectors.

In the context of human cognition, \textit{incompatibility} denotes that a different ordering of a series of judgments will lead to a different result, a phenomenon called the \textit{order effect}. Classical probability assumes that the measurements are always compatible and cannot capture such disturbances. The mathematical form of QP is a generalization of classical probability theory, which allows both compatible and incompatible measurements.

\subsection{Interference and Entanglement}
\label{sec:interence_entanglement}

\begin{concept}
(\textbf{Quantum Interference}).
Quantum interference refers to the phenomenon that two or more waves superimpose when they overlap in space, thus forming a new waveform~\cite{Ficek2005}. The two waves combine by adding their displacements together at every point in space and time to form a resultant wave of greater, lower, or the same amplitude, depending on the interference term. In quantum physics, the double-slit experiment proves that microscope particles have the wave-particle duality. \footnote{https://en.wikipedia.org/wiki/Wave\_interference}
\end{concept}

In the double-slit experiment, two slits are open for electrons to pass through to reach a detection screen. The two paths interfering with each other affect the position distribution of the particles on the detection screen and form an interference pattern that does not follow the classical law of total probability (LTP). This phenomenon cannot be explained sufficiently by classical theory, in which the LTP holds. Instead, the wave function $\varphi(x)$ is used to interpret this behavior. The wave function represents the probability amplitude of a particle at a position $x$ on the screen, and its square represents the probability. The state of the particle is in an interfered state affected by path 1 and path 2. In the multimodal sentiment analysis task, e.g., path 1 can be sentiment predicted based on the textual features of an utterance, and path 2 can be predicted based on visual features. The sentiment of the utterance is in an interfered state of two different modalities. This is similar to the state of an electron being in a \textit{quantum superposition} of the states of path 1 and path 2: 
\begin{equation} 
\varphi_p (x)= \alpha \varphi_1 (x) + \beta \varphi_2 (x) 
\end{equation}
where $x$ indicates the position after the particle reaches the detection screen,
 $\varphi_1 (x)$ and $\varphi_2 (x)$ are the wave functions of the two paths. $\alpha$ and $\beta$ are complex numbers satisfying $|\alpha|^2 + |\beta|^2 =1$. $|\alpha|^2$ and $|\beta|^2$ represent the probabilities of the electron passing through path 1 and path 2, respectively. The probability that the electron is at the state $\varphi_p$ can be calculated as: 
\begin{equation} 
	\begin{aligned}
		P (x)&=|\varphi_p(x)|^2=|\alpha\varphi_1(x)+\beta\varphi_2(x)|^2\\
		\quad\quad\,\,\,&=|\alpha\varphi_1(x)|^2+|\beta\varphi_2(x)|^2\\
		&+2|\alpha\beta\varphi_1(x)\varphi_2(x)|\cos{\phi},\\
	\end{aligned}
	\label{eq:inter2}
\end{equation}
where $\phi$ is the interference angle (i.e., relative phase between the two paths).  
\begin{equation} 
	\begin{aligned}
		I=2|\alpha\beta\varphi_1(x)\varphi_2(x)|\cos{\phi}
	\end{aligned}
\end{equation}
is the interference term that describes the interaction between the two paths. 
\begin{concept}
(\textbf{Quantum Entanglement})
\textit{Quantum entanglement} indicates a unique type of interaction between two or more particles in QT. Entangled particles are generated and measured, interact, or share spatial proximity such that the quantum state of each particle cannot be independently described. Particularly, operations on one particle cause the state of the other to change accordingly, even if they are far from each other. \footnote{https://en.wikipedia.org/wiki/Quantum\_entanglement}
\end{concept}

Mathematically, the joint state of the entangled particles, which lives in the tensor product of the Hilbert spaces of the particles, cannot be decomposed into a tensor product of the corresponding pure states in the individual Hilbert spaces. Assume two particles $A \in H_A$ and $B \in H_B$ are in an entangled state. The entanglement state $\ket{\psi_{AB}} \in H_A \otimes H_B$ cannot be expressed as a tensor product of states $\ket{\alpha_{A}} \in H_A$ and $\ket{\beta_{B}} \in H_B$.

\subsection{Probabilistic Interpretation of Quantum Theory}
\label{sec:probability}

QT delineates what a system is and how it evolves over time and establishes an entirely new set of rules governing what happens when systems are observed or measured. Since the establishment of QT, researchers have never stopped exploring the best method for interpreting quantum mechanics and statistical physics. Competing schools of thought give the same predictions in the regimes tested thus far. All of them lean on the idea of probability fundamentally. 

In QP, \textit{events} are subspaces of a Hilbert space, uniquely represented by the projector $P$ onto the space. An \textit{elementary event} is a 1-dimensional subspace $\ket{u}\bra{u}$, also called a \textit{dyad}. For a Hilbert space with a dimension greater than 2, Gleason's Theorem~\citep{Gleason1975} bridges QT and classical probability theory.

\begin{theorem}
\textbf{Gleason's Theorem}~\citep{Gleason1975}: There exists a one-to-one correspondence between a QP \textit{measure} $\mu(\cdot)$ and a density matrix $\rho$. For any event $P$, its QP defined by $\rho$ is given by $tr()$, which calculates the trace of a matrix:
\begin{equation}\label{eq:gleason}
    \mu_{\rho} (P) = tr(\rho P).
\end{equation}
\end{theorem}

In particular, for an elementary event $|u\rangle\langle u|$, $\mu_{\rho} (u) = \bra{u} \rho \ket{u} \in [0,1]$. It guarantees that $\sum_u \mu_{\rho} (\ket{u}\bra{u}) = 1$ for orthogonal basis events $\{ \ket{u}\bra{u} \}$. 

Classical probability theory is derived from the Kolmogorov axioms, e.g., nonnegativity, unitarity and additivity ~\cite{Kolmogorov2018}. Events in these axioms are defined as subsets of a universe sample space, which obey the Boolean logic where the distributive law holds. Therefore, the LTP can be derived from classical probability. In contrast, QP operates on subspaces as events. The distributive law is not necessarily obeyed, and the LTP can be absent. Therefore, QP can be seen as a generalization of classical probability. The relaxation of the distribution law makes QP capable of handling numerous paradoxes that classical probability usually yields in realistic decision-making scenarios~\citep{Yukalov2016}. This is also why quantum models have been applied in psychological and cognitive sciences~\citep{Ashtiani2015,Bagarello2012}. In the context of sentiment analysis, affective states such as sentiment, emotion and sarcasm are outputs of our cognitive process; thus, they can naturally be addressed with quantum-inspired models. This will be further described in the next section.

\textbf{\textit{Remark 1: Classical statistical physics vs. quantum mechanics.}}

It is known that the ontology of quantum mechanics is derived from the ontologies of classical physics and statistical physics. 
Therefore, they are closely related. 
Some similarities and differences 
~\footnote{
The central difference between statistical mechanics (equivalent to classical statistical mechanics) and quantum mechanics is that statistical ensembles are assigned to N particles in statistical mechanics, while in quantum mechanics such ensembles are assigned to one particle. Statistical mechanics is only a valid description 
for collections of many particles, while quantum mechanics can describe small systems; when these systems become larger, they can be described by quantum statistical mechanics and then by classical statistical mechanics as their size increases.} 
between quantum mechanics and statistical mechanics are given in Tab.~\ref{table-quantum-statistical} and Tab.~\ref{table-quantum-statistical-concepts} ~\citep{Kupervasser2009}.
Analogies between quantum mechanics and statistical mechanics in terms of mathematical aspects and physical foundations are briefly summarized in Tab.~\ref{table-quantum-statistical-statistical-nature} \citep{Abad2012}.

In ~\citep{Dirac1945}, Dirac designed a method to define general functions of non-commuting observables in quantum mechanics, which enables discussions in quantum mechanics about trajectories for the motion of a particle and also makes quantum mechanics more closely resemble classical mechanics.
~\citep{Abad2012} discussed the basic principles of classical statistical mechanics, especially those involving the concept of complementarity. Although there are significant differences between classical statistical mechanics and quantum mechanics in physics, the article argued that there are also some similarities between the two.
 For example, they both require some form of complementarity concept in describing the system state; e.g., Niels Bohr suggested that "the thermodynamical quantities of temperature and energy should be complementary in the same way as position and momentum in quantum mechanics" ~\citep{Bohr1972}. 
The article also mentioned that understanding complementarity can help us better understand the fundamental principles of statistical mechanics and provide some insights for a deeper understanding of quantum mechanics.

Some concepts or phenomena in quantum mechanics have classical counterparts. 
For example, it has been found in ~\cite{Pomorski2020} that quantum mechanical phenomena might be simulated by the classical statistical model.
In classical statistical physics or in epidemic modeling, the collapse of the probabilistic state vector is witnessed upon measurement. 
The classical density matrix can be derived from a quantum density matrix and described by the equation of motion in terms of the anticommutator.
In functional data processing, data can be encoded by orthogonal wave functions and the use of orthogonality, which can make things similar to quantum mechanics.

Another example is the quantum no-cloning theorem, which states that an unknown quantum state of a source system cannot be perfectly duplicated while leaving the state of the source system unperturbed ~\cite{Wootters1982, Dieks1982}. 
It was proven in ~\cite{Daffertshofer2002} that the classical cloning machine violates the Liouville dynamics governing the evolution of statistical ensembles. 
This kind of cloning process conflicts with the conservation of the Kullback-Leibler information distance ~\cite{Kullback1997} and the linearity of the Liouville dynamics. 

Pomorski conducted two studies ~\citep{Pomorski2020, Pomorski2023a} and verified the equivalence between a finite state stochastic machine, non-dissipative and dissipative tight-binding and the Schrödinger model as well as the equivalence between the classical epidemic model and non-dissipative and dissipative tight-binding models. 
Additionally, the authors suggested that these equivalence relationships can be used to study artificial life, quantum computing, and other fields, providing a basis for further research in these areas.

~\citep{uehara1986analogies} discussed the analogies between nonequilibrium classical statistical mechanics and quantum mechanics at the level of the Liouville equation and at the kinetic level and showed an analogy between the plasma kinetic theory and Bohm's quantum theory with `hidden variables'.
	
Classical mechanics and quantum mechanics follow similar mathematical principles, such as Poisson brackets and quantum commutation relations. 
In quantum mechanics, the observables (such as position and momentum) do not necessarily commute, and the order of their measurements affects the outcome. 
The concept of non-commuting observables is fundamental to the theory of QM and applies to any observables present in quantum mechanics. 
This non-commutativity rule is already present in theoretical mechanics as given by Poisson brackets,
meaning that the Poisson bracket rule is equivalent to the commutator rule known in quantum mechanics, as seen in the Heisenberg uncertainty principle ~\citep{Hamilton1833}.

However, calculations in these examples ~\citep{Pomorski2020,Wootters1982, Dieks1982,Daffertshofer2002,Kullback1997,Pomorski2020, Pomorski2023a,uehara1986analogies,Hamilton1833} are limited by strong constraints. Even though these phenomena can be explained to some extent by classical statistical physics, their applicability is still limited due to the strong constraints. As a more generalized theory, quantum physics can explain the aforementioned phenomena more understandably. Unlike the above works,
which need to find proofs and derive formulas under different aspects of physics (e.g., thermodynamics and statistical physics), these phenomena can be explained together under a unified quantum physical theory.

Probabilities arise from classical statistical mechanics due to a lack of knowledge about the target system. This limitation comes from the fact that the target system consists of a great number of particles\footnote{In quantum mechanics, the statistical ensemble is associated with each particle, while in statistical physics, the statistical ensemble is associated with a conglomerate of N particles.}, and it may be impossible to describe each one in detail. Therefore, statistical methods (e.g., statistical physics) are used to describe the entire system, and statistics are used to depict the properties of the target system from a macroscopic perspective. Once we know the detailed properties of every particle, then statistical physics is no longer needed~\cite{VanKampen1992, Abad2012}, and by using quantum mechanics, the properties of a particle can be efficiently described.
Thus, there are large differences between statistical mechanics and quantum mechanics ~\citep{Kupervasser2009}.
However, even though the particle properties can be thoroughly studied with quantum mechanics, the measurement result of the system still will not have a predictable outcome.
What we obtain is the probability of obtaining a certain outcome. 
This feature makes quantum mechanics more feasible for studying inherently uncertain human cognition processes.

\begin{table}[]
\begin{center}
\caption{Basic properties of quantum mechanics and statistical mechanics. This table is reproduced from Table 1 in ~\cite{Kupervasser2009}. }
\label{table-quantum-statistical}
\scalebox{0.66}{
\begin{tabular}{c|c}
\hline
\textbf{Quantum Mechanics}                                                                                                                                        & \textbf{Statistical Mechanics}                                                                                                                                                          \\ \hline
Density matrix                                                                                                                                           & Phase density function                                                                                                                                                         \\ \hline
Equation of motion for the density matrix                                                                                                                & Liouville equation                                                                                                                                                             \\ \hline
Wave packet reduction                                                                                                                                    & \begin{tabular}[c]{@{}c@{}}Coarsening of the phase density function or \\ the molecular chaos hypothesis\end{tabular}                                                          \\ \hline
\begin{tabular}[c]{@{}c@{}}Unavoidable interaction of the measured system with \\ the observer or the environment as described by reduction\end{tabular} & \begin{tabular}[c]{@{}c@{}}Theoretically infinitesimal but, in reality, a finite small interaction of \\ the measured system with the observer or the environment\end{tabular} \\ \hline
Nonzero and nondiagonal elements of the density matrix                                                                                                   & \begin{tabular}[c]{@{}c@{}}Correlations between the velocities and positions of\\ particles in different parts of the system\end{tabular}                                      \\ \hline
Pointer states                                                                                                                                           & \begin{tabular}[c]{@{}c@{}}Appropriate (i.e., macroscopically stable to small perturbations) \\ macroscopic states\end{tabular}                                                \\ \hline
\end{tabular}
}
\end{center}
\end{table}

\begin{table}[]
\begin{center}
\caption{Probability formulations in classical and quantum mechanics. This table is reproduced from Table 2 in ~\cite{Kupervasser2009}. }
\label{table-quantum-statistical-concepts} 
\scalebox{0.88}{
\begin{tabular}{c|c|c}
\hline
                        & \textbf{Statistical mechanics}          & \textbf{Quantum mechanics}           \\ \hline
Pure state              & Point $(q,p)$ of phase space & State vector     $\ket{\psi}$           \\ \hline
General state           & Probability density   $\rho(q,p)$         & Positive Hermitian operator $\rho$ \\ \hline
Normalization condition &      $\int\rho dqdp$                          &             $tr(\rho)=1$                \\ \hline
Condition for pure state &           $\rho=\delta-function$                     &          $\rho=\ket{\psi}\bra{\psi}$ (operator $\rho$ rank is equal to $1$)                  \\ \hline
Equation of motion      &       $\frac{\partial\rho}{\partial t}=\{H,\rho\}$       &         $i\hbar\frac{\partial\rho}{\partial t}=[H,\rho]$                    \\ \hline
Observable              & Function      $A(q,p)$                 & Hermitian operator    $A$      \\ \hline
Average value           &              $\int A\rho dqdp$                  &           $tr(A\rho)$                  \\ \hline
\end{tabular}
}
\end{center}
\end{table}

\begin{table}[]
\begin{center}
\caption{Comparison between quantum mechanics and classical statistical mechanics. Despite their different physical relevance, these theories exhibit several analogies as a consequence of their statistical nature. Classical statistical mechanics is represented by a Cognitive Ikegami Braitenberg vehicle with a sensing system \citep{Braitenberg1986,Iizuka2004}. This table is reproduced from Table 1 in ~\cite{Abad2012}. }
\label{table-quantum-statistical-statistical-nature} 
\scalebox{0.72}{
\begin{tabular}{c|c|c}
\hline
\textbf{Criterium}                                                                       & \textbf{Quantum mechanics}                                                       & \textbf{Statistical Mechanics}                                             \\ \hline
Parametrization                                                                 & Space-time coordinates  $(q,t)$                             & Mechanical macroscopic observables $I$                        \\ \hline
Probabilistic description                                                       & \begin{tabular}[c]{@{}c@{}}The wave function $\psi(q,t)$\\ $dp(q,t)=|\psi(q,t)|^2dq$ \end{tabular}                & \begin{tabular}[c]{@{}c@{}}The probability amplitude $\phi(I|\theta)$\\ $dp(I|\theta)=\phi^2(I|\theta)dI$\end{tabular}   \\ \hline
Deterministic theory                                                            & \begin{tabular}[c]{@{}c@{}}Classical mechanics in the limit\\ $\hbar\rightarrow0$\end{tabular} & \begin{tabular}[c]{@{}c@{}}Thermodynamics in the limit\\ $k\rightarrow0$\end{tabular} \\ \hline
\begin{tabular}[c]{@{}c@{}}Relevant physical \\ hypothesis\end{tabular}         & \begin{tabular}[c]{@{}c@{}}Correspondence principle: $\psi(q,t)\sim e^{\frac{i}{\hbar}S(q,t)}$\\ when $\hbar\rightarrow0$, where $S(q,t)$ is the action\end{tabular} & \begin{tabular}[c]{@{}c@{}}Einstein postulate: $\phi(I|\theta)\sim e^{\frac{1}{2k}S(I|\theta)}$\\ when $k\rightarrow0$, where $S(I|\theta)$ is the entropy\end{tabular} \\ \hline
Evolution                                                                       & Dynamical conservation laws                                             & Tendency towards thermodynamic equilibrium                        \\ \hline
Conjugated variables                                                            & \begin{tabular}[c]{@{}c@{}}Energy $E=\partial S(q,t)/\partial t$ momentum\\ $p=\partial S(q,t)/\partial q$\end{tabular}                                                                  & \begin{tabular}[c]{@{}c@{}}Restituting generalized forces\\ $\eta=\partial S(I|\theta)/\partial I$\end{tabular}                                    \\ \hline
Complementary quantities                                                        &                     $(q,t)\ versus\  (p,E)$                                                    &                    $I \ versus \ \eta$                                               \\ \hline
Operator representation                                                         & $\hat{q}^i = q^i$ and   $\hat{p}_i=-i\hbar\frac{\partial}{\partial q^i}$                                                           &  $\hat{I}^i=I^i$ and  $\hat{\eta}^i=2k\frac{\partial}{\partial I^i}$                                                              \\ \hline
\begin{tabular}[c]{@{}c@{}}Commutation \& \\ uncertainty relations\end{tabular} &                      $[\hat{q}^i,\hat{p}_j]=i\hbar\delta^{i}_{j}\Rightarrow\Delta q^i\Delta p_i 	\geq \hbar/2$                                                   &                                  $[\hat{I}^i,\hat{\eta}_j]=-2k\delta_{j}^{i}\Rightarrow\Delta I^i\Delta\eta_i\geq k$                                 \\ \hline
\end{tabular}
}
\end{center}
\end{table}

\section{A Quantum-Cognitive Perspective of Sentiment Analysis}
\subsection{Quantum Theory for Human Cognition}
As discussed in the previous sections, QP has been established as a more general formalism than classical probability theory and can better describe and explain a range of human cognitive processes and emotional activities. For example, in ~\cite{Pothos2009}, a QP model based on a Hilbert space representation and Schrodinger's equation is used to explain the violation of classical probability theory in the decision processes of two-stage gambling and the Prisoner's Dilemma games. To better describe decision vagueness, in ~\cite{Blutner2013}, the phenomenon of borderline contradictions is explained as a manifestation of quantum interference. Additionally, QP has been used to address human cognition and decision-making phenomena that are considered paradoxical, generate nonreductive understandings of human conceptual processing, and provide new understandings of perception and human memory~\citep{Bruza2015,Busemeyer2012}. 

User study is also a common method for verifying nonclassical phenomena. For example, ~\cite{Fell2019} presented quantum-inspired protocols based on the Stern-Gerlach experiment and the discrete Wigner function.

\begin{concept}[\textbf{Stern-Gerlach experiment}]
The Stern-Gerlach experiment demonstrated that the spatial orientation of angular momentum is quantized into two components (up and down). As such, an atomic-scale system was shown to have intrinsic quantum properties. ~\footnote{See more details in https://en.wikipedia.org/wiki/Stern\_Gerlach\_experiment}
\end{concept} 

\begin{concept}[\textbf{Wigner function}] The Wigner function provides a simple representation of the quantum state of a continuous system in the phase space and a method for performing quantum physics using probability distributions that can go negative ~\cite{DeBrota2020, Bianucci2002, Hillery1984}.
\end{concept}

The Stern-Gerlach experimental protocol has been used to exploit the quantum-cognitive structure for modeling users' cognitive states and the violation of certain classical probability axioms in users' multidimensional relevance judgments about documents. The results also show that QP theory is a better alternative to model multidimensional relevance judgments than its classical counterpart, i.e., Bayesian model~\cite{Uprety2020a}.

In another work~\cite{Uprety2019}, a user study was conducted by constructing the cognitive analog of an experiment in quantum physics to explain incompatibility and interference between multiple decision criteria in document relevance judgment. In ~\cite{Wang2016}, a user study was performed to explore and model the cognitive interference in users' relevance judgment about documents caused by the order effect of documents being judged.  

In classical probabilistic models such as Bayesian networks and connectionist networks, at every specific moment, a cognitive state is assumed to be in a definite state corresponding to a judgment or decision to be made. In contrast, quantum cognition stipulates that the system be in an indefinite state at each moment before the system is measured. Being in an indefinite state is similar to a system being in different basis states simultaneously. This is also called a superposition state~\citep{Nielsen2010}. Once the system is measured, the superposition state collapses onto one specific state, which is when the decision is made. 

In line with Bohr's theory~\citep{Bohr2010}, a cognitive system is in an indefinite state, and the measurement (e.g., asking a question) outcomes from the system are constructed by interaction of the indefinite state and the question being asked. The measurement (question) leads the superposition state to collapse onto a definite state. It is also argued that judgments are created in context rather than being prerecorded and extracted from memory~\citep{Payne1992}. Thus, compared with the assumption that the answer to a question is simply a reflection of an existing state, the QT-view better fits the psychological intuition for complex judgments~\citep{Busemeyer2012}.

Classical probability theory assumes the compatibility of events and obeys the LTP. However, decision-making experiments in psychology and cognitive science show a violation of these axioms analogous to the double-slit experiments in physics. QP, as a generalization of classic probability, allows superposition and incompatibility, and thus, the distributive axiom and the LTP are not necessarily obeyed. Therefore, QP-based models can provide a better modeling capability for our cognitive processes, which underpin sentiment and emotion analysis tasks. 

\subsection{Quantum Theory for Sentiment Analysis}
\begin{figure}[h]
	\includegraphics[width=0.6\textwidth]{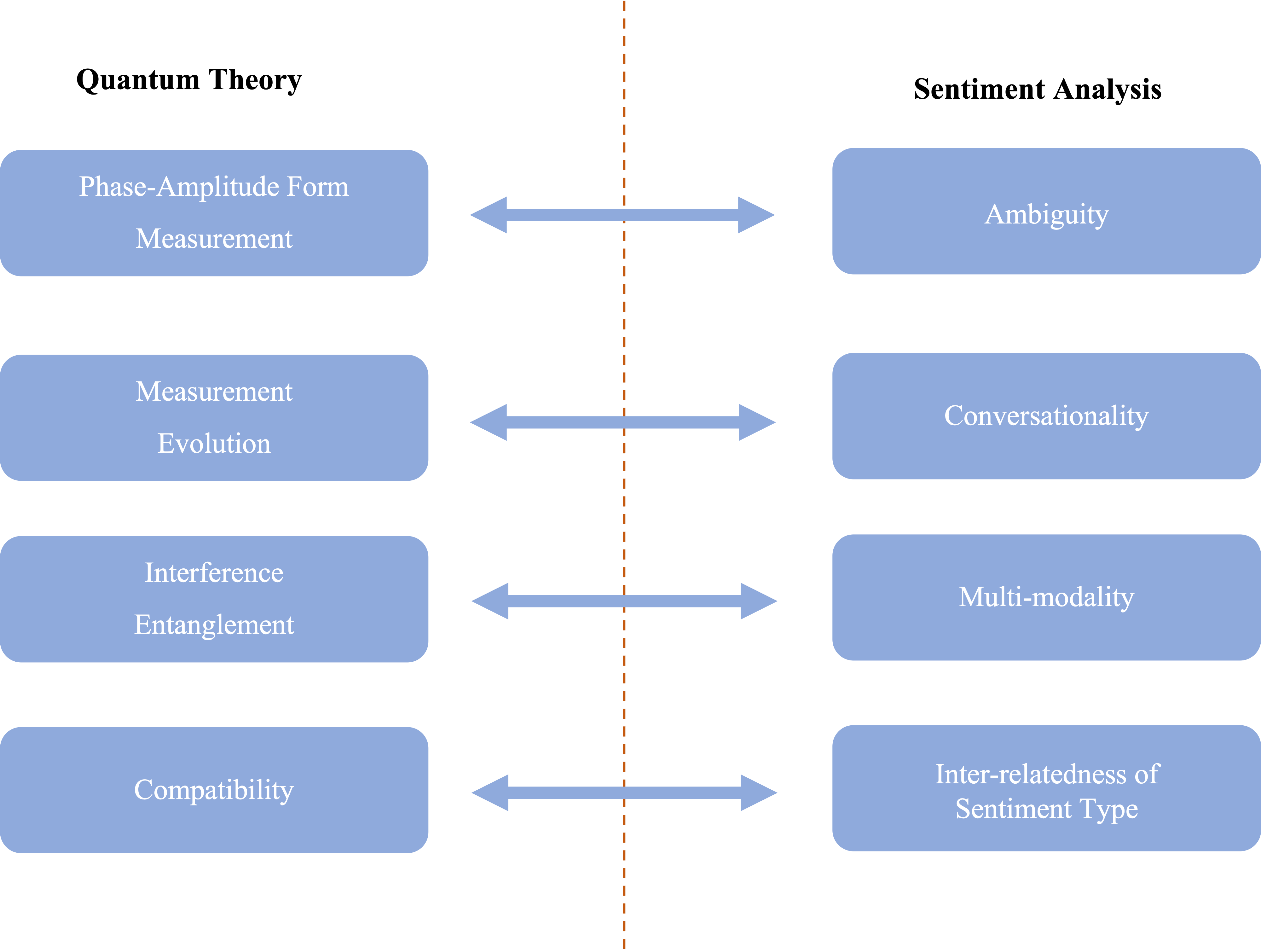}
	\caption{The conceptual mappings from Quantum Theory to Sentiment Analysis.}
	\label{fig:diagram-2}
\end{figure}

The above discussion illustrates why QT and QP are suitable for modeling human decisions and judgments and their advantages over classical probability theory. Since sentiment analysis is a typical human cognitive task, the idea of applying QT for sentiment analysis comes naturally. Beyond that, the quantum-theoretical framework is potentially fit for addressing the main challenges in sentiment analysis, as summarized in Fig.~\ref{fig:diagram-2}.\\

\noindent \textbf{Inherent Ambiguity in Sentiment Judgment.} One crucial challenge for sentiment analysis is the inherent ambiguity in the subjective judgment of sentiment. Quite often, a sentence contains different sentiment polarities toward different objects or from different perspectives. In the quantum probabilistic frameworks of cognitive modeling, the amplitudes and phases of complex representations are often used for encoding different aspects of semantics~\cite{Surov2021}. This makes QT a natural fit for encoding sentiment ambiguities into sentence representations: a document can be represented as a complex vector in which semantics can be associated with amplitudes, and the subjective sentiment is attributed to phases. Each vector dimension corresponds to a semantic aspect, and its phase indicates the sentiment toward the aspect. The overall sentiment of the sentence is determined by passing the complex vector to a measurement component to formulate the human judgment by aggregating the sentiments toward different aspects. The inherent uncertainty in the measurement outcomes can formulate the ambiguity well in the judgment process. 

It is worth noting that the subjective sentiment in the representation does not influence the semantic information but plays a role in the final sentiment judgment. Assume $z(x)=re^{i\theta}$ is the sentence representation on a single dimension; here, $x$ denotes a sentence in its "original" representation. When $\theta$ is given different values, e.g., $\theta_1=1/2\pi$ and $\theta_2=-1/4\pi$, it indicates different sentiments for the aspect, but the semantics given by $p(x)=|z(x)|^2=r^2$ remain unchanged. Nevertheless, it contributes to different probability distributions in the outcomes of the overall sentiment judgment.\\

\noindent \textbf{Conversationality.} The sentiment expressed by an utterance in a conversation is largely influenced by its conversational context. Under different contexts, a single utterance can take on different and even opposite sentiment polarities. This poses a crucial challenge on incorporating the conversational context in sentiment analysis, which often changes as the conversation proceeds. In QT, the concepts of \textit{measurement} and \textit{evolution} may be borrowed to render the conversational aspect of sentiment analysis. The configurations of a quantum measurement can affect the measurement outcome. As a simple example, when the spin of an electron is measured under different sets of bases, it leads to different outcomes. Analogously, the conversational context serves as the configuration of sentiment measurement performed by human perception. Therefore, quantum measurement can be employed to model the impact of conversational contexts on sentiment analysis. In addition, modeling the conversational context may be approached by the concept of quantum evolution to capture its evolving nature.\\

\noindent \textbf{Multimodality.} Human languages and communications are multimodal in nature. 
A sentiment is usually expressed multimodally. The speaker can express his or her sentiment via speech, gestures, eye contact, iris status, voice, etc. Through multimodal representation, the speaker's sentiment can be fully displayed and help the listener to better understand the speaker's intention. 
Different quantum concepts can account for the complicated interrelatedness between multimodal features in sentiment judgment. First, quantum interference offers inspiration for nonlinear multimodal fusion. The double-slit experiment tells us that the two paths interfering with each other affect the probability distribution of the particle reaching the final position on the detection screen. This is similar to the human cognition process in which textual and visual information disturb each other and affect the listener's understanding of the speaker's attitude. Assume $z_1(x_1)$ and $z_2(x_2)$ represent the probability amplitudes of modalities $x_1$ and $x_2$. Then, the multimodal probability amplitude is $z_3(x)=\alpha z_1(x_1)+\beta z_2(x_2)$, and $\alpha$ and $\beta$ are coefficients. The probabilities of $x_1$ and $x_2$ are their modulus squares $p_1(x_1)=|\alpha|^2|z_1(x_1)|^2$, $p_2(x_2)=|\beta|^2|z_2(x_2)|^2$. According to Equation~\ref{eq:inter2}, the multimodal probability is:
\begin{equation} 
	\begin{aligned}
    p_3(x_3)&=|z_3(x_3)|^2=|\alpha z_1(x_1)+\beta z_2(x_2)|^2 \\
    &=p_1(x_1)+p_2(x_2)+2\sqrt{p_1(x_1)p_2(x_2)}\cos{\theta} \\
    &=p_1(x_1)+p_2(x_2)+\sqrt{p_1(x_1)p_2(x_2)}(e^{i\theta}+e^{-i\theta})
	\end{aligned}
	\label{eq:multimodal_probability}
\end{equation}
Accordingly, the multimodal probability is a nonlinear fusion of the individual unimodal probabilities, where $\sqrt{p_1(x_1)p_2(x_2)}(e^{i\theta}+e^{-i\theta})$ is the interference item. This kind of fusion provides a higher level of abstraction \cite{Li2021,Jiang2020}. By leveraging quantum interference, three crucial tasks in sentiment analysis, i.e., multimodal representation learning, multimodal decision-making and multimodal information fusion problems can be addressed.

Moreover, quantum entanglement offers a nonseparable method for modeling the interactions of different modalities. In cognitive science, reductionism has been a powerful background philosophy underlying model development. In other words, a phenomenon can be analyzed by considering its different components separately, and the final result is synthesized by the result of each component. However, not every phenomenon is decomposable. For example, 
consider the sentiment of the utterance "Well, what a surprise!" and its corresponding audio. Judging from each modality (i.e., only text or audio is under consideration) without its context, the sentiment expression is ambiguous. Nevertheless, once its full context is utilized, the sentiment state of a particular sentiment judgment becomes apparent. This implies that we cannot consider distinct modalities individually. Quantum entanglement is a promising theory for modeling such nonseparability. In conjunction with the previously mentioned quantum interference, explainable and effective quantum-inspired methods are expected for handling multimodality issues in sentiment analysis.\\

\noindent \textbf{Interrelatedness of Different Sentiment Aspects.} As we stress in the beginning of the paper, sentiment is used as a broad notion covering a range of aspects, such as sentiment, emotion and sarcasm. Therefore, different sentiment aspects may coexist for a single sentence, and their judgments are often influenced by one another. 
The sentence "Silence is golden. Duct tape is silver." is a declarative sentence expressing a natural sentiment. However, in regard to a context in which the speaker is impatient with others, the sentence might express a sarcastic emotion with a negative sentiment to show the speaker's emotion of unhappiness. Thus, the judgment of sarcasm may depend on the sentiment analysis. 
Classical probabilistic frameworks may not address such a situation well, which assume that all events can be described within a single sample space under the so-called \textit{unicity} assumption. However, QT relaxes the assumption by allowing for incompatible events that do not share a common basis. These events cannot be combined into one event but must be evaluated sequentially, as the order of evaluation can influence the final result (i.e., the order effect). This is also often faced when judging multiple sentiment analysis aspects: recognizing one sentiment aspect may potentially affect the understanding of another. Incompatibility is key in distinguishing QP from classic probability and a crucial component for capturing the interrelatedness of sentiment aspects as multiple tasks. According to~\cite{Liu2021}, quantum incompatible measurement can describe correlations across tasks; thus, it is suitable for multitask decision-making problems. In addition, the concept of quantum relative entropy provides a metric to measure the degree of incompatibility between tasks.

\section{Quantum-Cognitively Inspired Sentiment Analysis Models}

Following the conceptual mappings between QT concepts and key challenges of sentiment analysis (Fig. 1), in this section, we discuss how each key challenge has been addressed by the quantum-cognitively inspired models. Due to the essential role of complex-valued representation of states in QT to capture the uncertainty of quantum systems. Such representation has also laid a theoretical foundation for quantum-cognitive sentiment analysis models. Thus, we first present an overview of complex-valued Hilbert space representation, which is compatible with QT and addresses the \textit{ambiguity} issue in semantic and sentiment representation. Then, we will go through the quantum-inspired sentiment analysis models that address the challenges of \textit{conversationality} (for conversational sentiment analysis), \textit{multimodality} (for multimodal sentiment analysis), the \textit{interaction of conversationality and multimodality} (for multimodal conversational sentiment analysis), and the \textit{interrelatedness of sentiment aspects} (for multitask multimodal conversational sentiment analysis). The oriented tasks and targeted key problems, as well as the adopted QT concepts of the surveyed models, are listed in Table~\ref{table:paper} and Table~\ref{table:problemandconcept}. 
Table~\ref{table:catbySaSeEm} to Table~\ref{table:modality} show different categorizations of the surveyed models by the tasks they were developed for, types of framework, quantum concepts that they used, and modalities that they process, respectively. A summary of the reported experimental results of these models is shown in Table~\ref{table:result}.

\begin{table*}[th]
	\begin{center}
	\caption{List of surveyed papers categorized by setting and task.}
	\label{table:paper}
	\scalebox{0.85}{
			\begin{tabular}{l|llll}
				\toprule
				\textbf{Reference} & \textbf{Year} & \textbf{Model} & \textbf{Setting} & \textbf{Task} \\
				\midrule
				
				\cite{Zhang2019} & 2019 & QSR & Basic Model & Sentiment Analysis \\
				\hline
				
				\cite{Li2021a} &2021 & QI-CNN & Basic Model & Sentiment Analysis \\
				\hline
				
				\cite{Zhang2019a} & 2019 & QIN & Conversational Context & Sentiment Analysis \\
				\hline
				
				\cite{Guo2019} & 2019 & DMATT-BiGRU & Conversational Context & Sentiment Analysis \\
				\hline
				
				\cite{Zhang2021cfn} & 2021 & CFN & Conversational Context & Sarcasm Detection \\
				\hline
				
				\cite{Zhang2018} &2018 & QMSA & Multimodal Framework & Sentiment Analysis  \\
				\hline
				
				\cite{Gkoumas2021} & 2021 & DecisionFusion & Multimodal Framework & Sentiment Analysis  \\
				\hline
				
				\cite{Gkoumas2021a} &2021 & EFNN & Multimodal Framework & Sentiment Analysis  \\
				\hline
				
				\cite{Li2021} &2021 & QMF & Multimodal Framework & Sentiment Analysis  \\
				\hline
				
				\multirow{2}{*}{\cite{Zhang2020}} & \multirow{2}{*}{2020} & \multirow{2}{*}{QMN} & Conversational Context & \multirow{2}{*}{Sentiment Analysis} \\
				& & & Multimodal Framework & \\
				\hline
				
				\multirow{2}{*}{\cite{Li2021b}} & \multirow{2}{*}{2021} & \multirow{2}{*}{QMNN} & Conversational Context & \multirow{2}{*}{Emotion Recognition} \\
				& & & Multimodal Framework & \\
				\hline
				
				\multirow{3}{*}{\cite{Liu2021} } & \multirow{3}{*}{2021} & \multirow{3}{*}{QPM} & Multitask Learning & Sarcasm Detection  \\
				& & & Conversational Context &  \\
				& & & Multimodal Framework & Sentiment Analysis \\
				\bottomrule
			\end{tabular}}
	\end{center}
\end{table*}

\begin{table*}[th]
\begin{center}
\caption{List of key problems in sentiment analysis that each quantum-inspired model addresses and the corresponding QT concepts employed}
\label{table:problemandconcept}
\scalebox{0.85}{
\begin{tabular}{lll}
\toprule
\textbf{Model} & \textbf{Key Problems}                                                                                                                                                        & \textbf{Quantum Theory Concepts}                                                                                                                                                          \\ 
\midrule
QI-CNN            & \begin{tabular}[l]{@{}l@{}}Representation Model\\ \end{tabular}                                                                           & \begin{tabular}[l]{@{}l@{}}Density Matrix\end{tabular}                                                                                                             \\ \hline
QSR            & \begin{tabular}[l]{@{}l@{}}Representation Model\\ \end{tabular}                                                                           & \begin{tabular}[l]{@{}l@{}}Density Matrix\end{tabular}                                                                                                             \\ \hline
QIN            & \begin{tabular}[l]{@{}l@{}}Representation Model\\ Conversational Context Interaction\end{tabular}                                                                           & \begin{tabular}[l]{@{}l@{}}Density Matrix\\ Quantum Measurement\end{tabular}                                                                                                             \\ \hline
DMATT-BiGRU    & \begin{tabular}[l]{@{}l@{}}Representation Model\\ Conversational Context Interaction\end{tabular}                                                                           & Density Matrix                                                                                                                                                                           \\ \hline
CFN            & \begin{tabular}[l]{@{}l@{}}Representation Model\\ Conversational Context Interaction\end{tabular}                                                                           & \begin{tabular}[l]{@{}l@{}}Complex-Valued Representation\\ Density Matrix\\ Quantum Composite System\\ Quantum Measurement\end{tabular}                                                  \\ \hline
QMSA           & \begin{tabular}[l]{@{}l@{}}Multimodal Representation Model\\ Multimodal Information Interaction\end{tabular}                                                              & \begin{tabular}[l]{@{}l@{}}Density Matrix\\ Quantum Interference\end{tabular}                                                                                                            \\ \hline
DecisionFusion & Multimodal Decision Fusion                                                                                                                                                 & \begin{tabular}[l]{@{}l@{}}Complex-Valued Representation\\ Quantum Incompatibility\\ Quantum Measurement\end{tabular}                                                                    \\ \hline
EFNN           & \begin{tabular}[l]{@{}l@{}}Multimodal Representation Model\\ Multimodal Information Interaction\end{tabular}                                                              & \begin{tabular}[l]{@{}l@{}}Complex-Valued Representation\\ Quantum Evolution\\ Quantum Entanglement \\ Quantum Measurement\end{tabular}                                                      \\ \hline
QMF           & \begin{tabular}[l]{@{}l@{}}Multimodal Representation Model\\ Multimodal Information Interaction\end{tabular}                                                              & \begin{tabular}[l]{@{}l@{}}Complex-Valued Representation\\ Quantum Evolution\\ Quantum Entanglement \\ Quantum Measurement\end{tabular}                                                      \\ \hline
QMN            & \begin{tabular}[l]{@{}l@{}}Conversational Context Interaction\\ Multimodal Representation Model\\ Multimodal Information Fusion\end{tabular}                              & \begin{tabular}[l]{@{}l@{}}Density Matrix\\ Quantum Interference\\ Quantum Measurement\end{tabular}                                                                                      \\ \hline
QMNN           & \begin{tabular}[l]{@{}l@{}}Conversational Context Interaction\\ Multimodal Representation Model\\ Multimodal Information Fusion\end{tabular}                              & \begin{tabular}[l]{@{}l@{}}Complex-Valued Representation\\ Density Matrix\\ Quantum Evolution\\ Quantum Measurement\end{tabular}                                                         \\ \hline
QPM            & \begin{tabular}[l]{@{}l@{}}Conversational Context Interaction\\ Multimodal Representation Model\\ Multimodal Information Fusion\\ Multitask Decision-Making\end{tabular} & \begin{tabular}[l]{@{}l@{}}Complex-Valued Representation\\ Density Matrix\\ Quantum Composite System\\ Quantum Interference\\ Quantum Incompatibility\\ Quantum Measurement\end{tabular} \\
\bottomrule
\end{tabular}}
\end{center}
\end{table*}

\begin{table*}[th]
\begin{center}
\caption{Categorization of surveyed models by task.}
\label{table:catbySaSeEm}
\scalebox{0.85}{
\begin{tabular}{ll}
\toprule
\textbf{Task} & \textbf{Model} \\ 
\midrule
Sentiment Analysis         & \begin{tabular}[l]{@{}l@{}}QSR\cite{Zhang2019} \\ QI-CNN\cite{Li2021a} \\ QIN\cite{Zhang2019a} \\ DMATT-BiGRU\cite{Guo2019} \\ QMSA\cite{Zhang2018} \\ DesicionFusion\cite{Gkoumas2021} \\ EFNN\cite{Gkoumas2021a} \\ QMF\cite{Li2021} \\ QMN\cite{Zhang2020} \\ QPM\cite{Liu2021}  \end{tabular}            \\ \hline
Sarcasm Detection            & \begin{tabular}[l]{@{}l@{}}CFN\cite{Zhang2021cfn}\\ QPM\cite{Liu2021} \end{tabular}   \\ \hline
Emotion Recognition            & \begin{tabular}[l]{@{}l@{}}QMNN\cite{Li2021b}\end{tabular}\\
\bottomrule
\end{tabular}}
\end{center}
\end{table*}

\begin{table*}[th]
\begin{center}
\caption{Categorization of surveyed models by setting.}
\label{table:setting}
\scalebox{0.85}{
\begin{tabular}{ll}
\toprule
\textbf{Task} & \textbf{Model} \\ 
\midrule
Basic Model         & \begin{tabular}[l]{@{}l@{}}QSR\cite{Zhang2019} \\ QI-CNN\cite{Li2021a} \end{tabular}            \\ \hline
Conversational Context Model& \begin{tabular}[l]{@{}l@{}}QIN\cite{Zhang2019a} \\ DMATT-BiGRU\cite{Guo2019} \\ CFN\cite{Zhang2021cfn}\\ QMN\cite{Zhang2020} \\ QMNN\cite{Li2021b} \\ QPM\cite{Liu2021} \end{tabular}   \\ \hline
Multimodal Framework & \begin{tabular}[l]{@{}l@{}}QMSA\cite{Zhang2018} \\ DesicionFusion\cite{Gkoumas2021} \\ EFNN\cite{Gkoumas2021a} \\  QMF\cite{Li2021} \\ QMN\cite{Zhang2020} \\ QMNN\cite{Li2021b} \\ QPM\cite{Liu2021} \end{tabular}\\ \hline
Multitask Learning & \begin{tabular}[l]{@{}l@{}}QPM\cite{Liu2021}  \end{tabular}\\
\bottomrule
\end{tabular}}
\end{center}
\end{table*}

\begin{table*}[th]
\begin{center}
\caption{Categorization of surveyed models by quantum concept.}
\label{table:concept}
\scalebox{0.85}{
\begin{tabular}{ll}
\toprule
\textbf{Task} & \textbf{Model} \\ 
\midrule
Density Matrix         & \begin{tabular}[l]{@{}l@{}}QSR\cite{Zhang2019} \\ QI-CNN\cite{Li2021a} \\ QIN\cite{Zhang2019a} \\ DMATT-BiGRU\cite{Guo2019} \\ CFN\cite{Zhang2021cfn} \\ QMSA\cite{Zhang2018} \\ QMN\cite{Zhang2020} \\ QMNN\cite{Li2021b} \\ QPM\cite{Liu2021} \end{tabular}            \\ \hline
Quantum Measurement& \begin{tabular}[l]{@{}l@{}}QIN\cite{Zhang2019a} \\ CFN\cite{Zhang2021cfn} \\ DesicionFusion\cite{Gkoumas2021} \\ EFNN\cite{Gkoumas2021a} \\ QMF\cite{Li2021} \\ QMN\cite{Zhang2020} \\ QMNN\cite{Li2021b} \\ QPM\cite{Liu2021} \end{tabular}   \\ \hline
Complex-Valued Representation         & \begin{tabular}[l]{@{}l@{}}CFN\cite{Zhang2021cfn} \\ DesicionFusion\cite{Gkoumas2021} \\ EFNN\cite{Gkoumas2021a} \\ QMF\cite{Li2021} \\ QMNN\cite{Li2021b} \\ QPM\cite{Liu2021}  \end{tabular}     \\ \hline
Quantum Composite System   & \begin{tabular}[l]{@{}l@{}}CFN\cite{Zhang2021cfn} \\ QPM\cite{Liu2021}  \end{tabular}            \\ \hline
Quantum Interference         & \begin{tabular}[l]{@{}l@{}}QMN\cite{Zhang2020} \\ QPM\cite{Liu2021} \end{tabular}    \\ \hline
Quantum Incompatibility    & \begin{tabular}[l]{@{}l@{}}DesicionFusion\cite{Gkoumas2021} \\ QPM\cite{Liu2021} \end{tabular}   \\ \hline
Quantum Evolution         & \begin{tabular}[l]{@{}l@{}} EFNN\cite{Gkoumas2021a} \\ QMF\cite{Li2021} \\ QMNN\cite{Li2021b} \end{tabular}            \\ \hline
Quantum Entanglement& \begin{tabular}[l]{@{}l@{}}EFNN\cite{Gkoumas2021a} \\ QMF\cite{Li2021}  \end{tabular}\\
\bottomrule
\end{tabular}}
\end{center}
\end{table*}

\begin{table*}[th]
\begin{center}
\caption{Categorization of surveyed models by modality}
\label{table:modality}
\scalebox{0.85}{
\begin{tabular}{ll}
\toprule
\textbf{Task} & \textbf{Model} \\ 
\midrule
Textual Input & \begin{tabular}[l]{@{}l@{}}QSR\cite{Zhang2019} \\ QI-CNN\cite{Li2021a} \\ QIN\cite{Zhang2019a} \\ DMATT-BiGRU\cite{Guo2019} \\ CFN\cite{Zhang2021cfn} \\ QMSA\cite{Zhang2018} \\ DesicionFusion\cite{Gkoumas2021}\\ EFNN\cite{Gkoumas2021a}  \\ QMF\cite{Li2021} \\ QMN\cite{Zhang2020} \\ QMNN\cite{Li2021b} \\ QPM\cite{Liu2021} \end{tabular}            \\ \hline
Visual Input & \begin{tabular}[l]{@{}l@{}}  QMSA\cite{Zhang2018} \\ DesicionFusion\cite{Gkoumas2021}\\  EFNN\cite{Gkoumas2021a} \\ QMF\cite{Li2021} \\ QMN\cite{Zhang2020} \\ QMNN\cite{Li2021b} \\ QPM\cite{Liu2021} \end{tabular}   \\ \hline
Acoustic Input         & \begin{tabular}[l]{@{}l@{}} DesicionFusion\cite{Gkoumas2021} \\ EFNN\cite{Gkoumas2021a} \\ QMF\cite{Li2021} \\ QMNN\cite{Li2021b} \end{tabular}    \\
\bottomrule
\end{tabular}}
\end{center}
\end{table*}

\subsection{Addressing Ambiguity: Complex Hilbert Space Representation of Semantics and Sentiment}

The complex Hilbert space representation was first adopted to model information semantics and, more recently, used as a unified representation model to integrate semantics and sentiment. 

\subsubsection{Complex Hilbert Space Representation of Semantics}
By making an analogy to particles with different states disturbing each other and giving rise to new states depending upon their relative phases,~\cite{Li2018} used the Hilbert space representation of words to derive the meaning of a combination of words. In their model, a word is a linear combination of latent concepts with complex weights, and the combination of words is viewed as a complex combination of word states, either a superposition state or a mixed state. 
Two approaches were proposed to compute the density matrices representing sentences, namely, the complex embedding superposition (CE-Sup) network and complex embedding mixture (CE-Mix) network. The models performed better than various nonquantum baseline models on the binary sentence classification task.

Similarly,~\cite{Wang2019} posited that semantic linearity does not always hold in human language understanding and treats \textit{sememes} (i.e., minimum semantic units in linguistics) as the basis of a Hilbert space. Words can be modeled as superposition states of these sememes. After determining the word representations, the concept of quantum mixture is leveraged to formulate each word composition as a mixed system composed of the word states represented by a density matrix. The role of the complex phases in word representations has also been exploited to implicitly entail the quantum interference between words. By applying this new representation, a quantum probability-driven neural network was designed for the text classification task. Evaluation results on six benchmarking datasets demonstrate the model effectiveness.

Targeting capturing the nonclassical correlations within the word sequences, ~\cite{Chen2020a} proposed a neural network model with a novel entanglement embedding module. The function of the module is to transform a word sequence into an entangled pure state representation. The entanglement embedding was implemented by a complex-valued neural network, which is essentially an approximation of the unitary operation that converts the initial product state to an entangled state. After entanglement embedding, quantum measurement is applied to extract high-level features of the word sequences. The proposed model was tested effectively on two benchmark question-answering (QA) datasets. Moreover, entanglement entropy was employed to explain the word interactions post hoc.

In summary, quantum-inspired models can lead to decent complex-valued semantic representations. Since numerical constraints on model layers need to be imposed to support the quantum formulation, the representation capacity of the model may drop as a consequence. However, many experiments show that the complex Hilbert space representation can achieve comparable or even stronger performance compared with traditional models, alleviating such worries. It provides stronger empirical support for applying quantum-theoretical frameworks to the sentiment analysis context.

\subsubsection{Encoding Sentiment in Quantum-inspired Hilbert Space Representation}

The most fundamental task for sentiment analysis is to build a unified representation that captures both semantics and sentiment information. Intuitively, the sentiment of a sentence (or utterance) adheres to the semantics it expresses. In quantum-inspired sentiment analysis models, the problem is encoding sentiment information into the aforementioned Hilbert space semantic representation as density matrices. 

\begin{shaded}
\textit{\textbf{Task Definition: Basic Sentiment Analysis}. 
Let $\{s^j, y^j \}$ be the $j^{th}$
sample in a dataset, where $s^j$ is a sentence, $y^j$ is its sentiment label, $j\in [1,N]$, and $N$ is the size of
the dataset. The objective is to establish a function, mapping $s^j$ to its corresponding sentiment label $y^j$. }
\end{shaded}

In sentiment analysis tasks, representation learning is a key problem. Inspired by the superiority of semantic Hilbert space representation (as density matrices), as introduced in the previous subsection, sentiment-aware density matrix sentence representations have been constructed.
For this purpose, in an early work \cite{Zhang2019}, a QSR model was designed. First, a hidden Markov model (HMM)-based part-of-speech tagger is used to extract phrases containing adjectives, adverbs and sentiment phrases. Then, the extracted items are modeled as projectors in a Hilbert space. The sentence representation, as a density matrix, is computed through the maximum likelihood estimation over the set of projectors. To smooth the density matrices, Dirichlet smoothing is also applied. The proposed representation model was tested with five classification methods on two datasets (OMD and Sentiment140) and showed a performance improvement.

\cite{Li2021a} proposed a quantum-inspired convolutional neural network (QI-CNN). It also leveraged the concept of the density matrix to encode the mixture of semantic subspaces and capture the feature interactions between sentiment words. Embedding vectors are treated as the observed state for each word. Thus, a sentence corresponds to the mixture state represented by a density matrix. Unlike the previous work~\cite{Zhang2019}, the density matrix here is not computed through the maximum likelihood estimation. Instead, it is integrated into a neural network architecture and automatically updated through back-propagation. The density matrix representing each sentence is then fed into a CNN to obtain the sentiment classification result.

However, in these models, complex values are not used, limiting their modeling capability. The concepts from QT are not fully explored for encoding multimodal information and contextual interactions, which are among the crucial challenges for sentiment analysis. Furthermore, the potential of quantum-inspired methods for addressing the interrelatedness of multiple tasks, such as sarcasm detection and emotion recognition, has not been investigated. 

More recent research has revealed the advantage of applying complex-valued Hilbert space representation in sentiment analysis, which is detailed in the following subsections. The complex-valued representation usually follows a polar decomposition of complex numbers, splitting the complex representation into \textit{amplitudes} and \textit{phases}. With the complex values, we can integrate the semantic and sentiment representations in a unified form, e.g., by encoding rich semantic information through amplitudes and sentiment information by phases. In practice, the amplitudes are often initialized with pretrained word embeddings, yet initializing the phases is still an open question. Complex values are processed to form the density matrix to represent a sentence. Then, according to different quantum-inspired components used to deal with downstream tasks, the density matrix is processed and trained differently, naturally forming a representation learning procedure. 

In summary, the complex-valued Hilbert space representation forms a well-principled and solid foundation for the recent quantum-inspired sentiment analysis models, based on which the other key challenges are addressed. The detailed model structures vary depending on the specific sentiment analysis tasks and the challenges addressed, which we introduce next.

\subsection{Addressing Conversationality: Quantum-inspired Conversational Sentiment Analysis}

\begin{shaded}
\textit{\textbf{Task Definition: Conversational Sentiment Analysis}. Typically, a dataset for conversational sentiment analysis has $L$ samples (conversations), where the $j^{th}$ sample is denoted as $\{X_j=(u_m), Y_j=(y_m)\}$, $m\in[1,n]$, $n$ is the number of utterances in a conversation, $u_m$ represents the $m^{th}$ utterance and $y_m$ is its sentiment label. Under this typical setup, the objective is to establish a function to map each utterance in a conversation to its corresponding sentiment label. Note that there exist two alternative task setups. First, after learning the contextual interaction, each conversation is treated as a whole. In this setup, $Y_j$ denotes the sentiment label for a single conversation. The objective then becomes establishing a function to map a conversation to its sentiment label. Second, the last utterance in a conversation is regarded as the target utterance that needs to be classified, and its preceding utterances are considered context. This setup means $X_i=(c_k,t), where k\in[1,n-1]$ and $t$ is the last utterance. This alternative studies the interactions with context and predicts the target utterance's sentiment polarity with the captured contextual information.}
\end{shaded}

Compared with the basic sentiment analysis task, a particular challenge for conversational sentiment analysis is to model the contextual interactions between utterances in a conversation, i.e., the issue of conversationality. Recurrent neural networks, e.g., RNN, LSTM and GRU, can capture contextual interactions. In quantum-inspired models, quantum evolution, as a theory describing the change in quantum states over time, has been used to analogize the evolution of a speaker's emotional state during conversation. In addition, the concept of a quantum composite system that describes the interaction between a quantum system and its surrounding environment has also been applied in conversational sentiment analysis models.

\cite{Zhang2019a} proposed the QIN model, which jointly captures the intrautterance and interutterance interactions through the mathematical formalism of QT and a LSTM structure. Textual features are first extracted and contextualized through a combination of density matrix representation and a convolutional neural network (DM-CNN). With DM-CNN, the interactions between words within an utterance are captured to compose the utterance-level representation, which is then used to extract strong and weak interutterance interaction patterns analogous to the strong-weak measurement formalism in QT. Finally, the two kinds of interactions are integrated into the LSTM output gates, and the LSTM hidden states are used for sentiment classification by a softmax function. By learning both the intra- and interutterance interaction dynamics, the QIN achieved the best performance over all baselines. For example, on the IEMOCAP dataset, compared with the contextual biLSTM model \cite{Poria2017}, QIN improves the performance by 7.1\% on accuracy.
The CNN-LSTM backbone structure endows the model with an ability for both sequence modeling and local feature extraction. Moreover, introducing quantum concepts succeeded in capturing the intra- and interutterance interactions. However, the number of LSTM components for each turn in the conversation is determined by the number of speakers, and it varies as the conversation continues. Note that if the  concern is computational efficiency and convergence speed, a gated recurrent unit (GRU) can be used instead of LSTM.

Inspired by the density matrix representation of a mixture system in quantum mechanics and statistical physics, ~\cite{Guo2019} designed a quantum-inspired model called DMATT-BiGRU, which combines a density matrix-based attention mechanism and a bidirectional GRU structure ~\cite{Guo2019}. They introduced the so-called static and dynamic density matrices to capture the intrautterance and interutterance interactions, which are input to the GRU and the attention layer, respectively. Different from other quantum-inspired sentiment analysis models, in this work, the probability distribution is not derived by quantum measurement but computed from the attention mechanism. Compared with baseline models including CNN \cite{Kim2014}, BiLSTM \cite{Poria2017} and ATT-BiLSTM \cite{Wang2016a}, on three datasets, DMATT-BiGRU obtained the best results, with improvements ranging from 1.1\% to 9.3\% on accuracy.
The main idea of this work is similar to the work in~\cite{Zhang2019a}, but their implementation details slightly differ. Although these approaches have demonstrated promising performance in effectiveness, their computational cost is relatively high compared with the normal attention mechanism. Moreover, their context modeling capability is limited; not only does the immediately preceding utterance have an impact on the current discourse but also all the historical utterances in the conversation can influence the current utterance. 

Quantum-inspired models have also been developed for sarcasm detection in a conversational context. Sarcasm is a manifestation of human cognition that embodies intrinsic uncertainties. ~\cite{Zhang2021cfn} proposed a CFN that leverages QP and fuzzy logic to capture the vagueness and uncertainty of human language in emotional expression and understanding. CFN is composed of three components. First, the target utterance and its conversational context are encoded in a single complex-valued density matrix, which captures the contextual interactions as a composite system via a fuzzy membership function. Second, a fuzzy measurement is applied to obtain sarcastic features. Finally, a softmax function is used for classification. The CFN model was tested on two sarcasm recognition benchmarking datasets and compared with 9 baseline models. It achieved an average improvement of 8.0\% in accuracy over multitask learning (MTL) \cite{Chauhan2020} on the MUStARD dataset and comparable performance with a state-of-the-art model RCNN-RoBERTa (which is based on the large-scale pretrained language model RoBERTa) \cite{Potamias2020} on the Reddit-track dataset.
Different from the two previously discussed works \citep{Zhang2019a, Guo2019} that build quantum-inspired components upon classical DL components, all the components in CFN are quantum-inspired. Using the concepts of QP and fuzzy theory, CFN addresses the spontaneous and hence inherently uncertain expression of human emotion from a quantum-cognitive perspective. For the first time, the complex-valued representation was used in the sarcasm detection task, which, according to the experimental results, made a remarkable contribution to the model's overall performance. However, computing quantum composite systems is complicated, and initializing the complex phase remains an open problem that calls for further exploration.

\subsection{Addressing Multimodality: Quantum-inspired Multimodal Sentiment Analysis}

In addition to leveraging contextual information as in the models discussed in the previous subsection, features from different modalities also provide plentiful sentimental hints. Modeling different modalities for sentiment analysis is a challenging problem. This is because the sentiment polarity (e.g., positive, neutral, or negative) that an utterance may reflect is influenced by other modalities that act as context. 

\begin{shaded}
\textit{\textbf{Task Definition: Multimodal sentiment analysis}. Given a dataset of size $L$, each utterance is composed of multiple modalities (e.g., any combinations of textual, visual, and acoustic features). In the most general setting, a sample can be represented as $\{X_j,Y_j\}$, $j\in[1,L]$, $X_j=(u^t, u^v, u^a)$, where $u^t, u^v, u^a$ represent textual, visual, and acoustic features, respectively, and $Y_j$ is the sentiment label. The objective is to establish a function that maps each utterance $X_j$ to its corresponding sentiment label $Y_j$. This task can be decomposed into several core subtasks. The fusion of multimodal information can be at the representation learning process or the decision-making stage, giving rise to the challenge of multimodality in terms of multimodal information fusion, multimodal representation learning and multimodal decision-making. The main QT mechanisms used to address this challenge include quantum entanglement to capture the nonseparability of different modalities in feature-level fusion and quantum interference to bridge the ``semantic gap'' between modalities and cognitive interference in decision fusion.}

\end{shaded}
\cite{Zhang2018} proposed a QMSA framework. It comprises two components: a QMR model and a multimodal decision fusion strategy inspired by quantum interference (QIMF). QMR represents textual and visual information in separate density matrices. Similar to the density matrix construction process in~\cite{Zhang2019}, QMR built density matrices through maximum likelihood estimation and used them for sentiment classification.
  
Furthermore, the QIMF drew an analogy between the interplay of sentiment decisions on two modalities and the quantum interference of electrons passing through two slits in the double-slit experiment. Accordingly, the formalism of quantum interference is employed to fuse multimodal decisions. In contrast to traditional additive or concatenative fusion, the quantum interference-inspired method provides a nonlinear fusion method. Experiments were conducted on the Flickr and Getty image datasets. QMSA in combination with the random forest (RF) classifier achieved the best result. When combined with the support vector machine (SVM) classifier, it obtained a comparable result to a strong baseline deep convolutional neural network (DCNN) \cite{Yu2016} model. On the Flickr dataset, QMSA obtained the best results with both classifiers.
 
Despite the improved performance, QMSA uses two separate models for representation and fusion, which are trained separately instead of end-to-end. Moreover, the absence of a complex-valued representation limits the model from capitalizing on the full power of QP. 

\cite{Gkoumas2021} introduced a novel decision-level fusion strategy inspired by quantum cognition. The task is to predict the sentiment of utterances in videos. It assumes that unimodal sentiment judgments do not occur independently. They are informatively interactive and influenced by other modalities that act as a context for the current modality. Utterances are first formulated as quantum superpositions of positive and negative sentimental polarities. Unimodal classifiers are built as mutually incompatible observables on a complex-valued Hilbert space, which is spanned by different unimodal sentiment basis vectors. The unimodal observables are estimated from the training data. Then, the final multimodal sentiment result is established from the learned unimodal observables by considering the interference between them. Experiments were conducted on the CMU-MOSI and CMU-MOSEI datasets. The proposed method outperformed a range of existing decision-level fusion approaches (e.g., stacking method \cite{PontiJr2011}) and content-level fusion approaches (e.g., QMF \cite{Li2021}).

Remarkably, this work was the first quantum-cognitively inspired approach to model the incompatibility between multimodal sentiment judgments in video sentiment analysis. Fundamentally different from the previous approaches, it takes the sentimental polarities of different modalities as the basis of incompatible subspaces representing the unimodal cognitive states. These subspaces are estimated from the statistics of the training data.

The models discussed above are focused on decision-level fusion, while the unimodal classification results are obtained individually. Therefore, the interactions between multimodal contents were not incorporated at the earlier stages. Indeed, multimodal information often acts nonseparably and jointly determines a sentiment decision, similar to the entanglement phenomenon in QT. 

Based on this motivation, \cite{Gkoumas2021a} proposed a quantum probabilistic neural model called the EFNN. EFNN takes multimodal inputs and feeds them into three separate bimodality fusion neural networks. The multimodal inputs are projected into a bidimensional space, and then the information is transformed into its quantum analog, the quantum state, through a preparation step. Then, through the tensor product operation between each pair of bimodal representations, a pairwise modality fusion (as a composite system) is generated. With the evolution of context (formally under Hamiltonian evolution), the composite system enters an entangled (i.e., nonseparable) state. A set of parameterized measurement operators then map the complex-valued representations to real-valued high-level representations via quantum measurement. Then, a row-max pooling operator is applied, and finally, the result from a fully connected layer is passed to the softmax function for classification.
EFNN was evaluated on the widely used datasets CMU-MOSI and CMU-MOSEI. Compared with a wide range of baseline models, EFNN obtained an average boost of 3\% over the next best models (c-GRU \cite{Ghosal2018} and QMF ~\cite{Li2021}).

Different from other studies that employed the concept of quantum entanglement as a theoretical explanation, EFNN operationalizes a feature fusion strategy by explicitly integrating the formalism of quantum entanglement in a neural network structure.
The Hamiltonian evolution creates nonseparable fusion for quantum composite systems. It is worth noting that this work also illustrated the advantages of quantum-inspired models from the perspective of post hoc interpretability by investigating the bimodal correlations within the composite utterance states. A case study showed that the most entangled pairs are those in which one of two modalities is ambiguous or uninformative.
In contrast, when the context of both modalities is sufficiently informative and unambiguous, entanglement does not exist. In this case, the composite representation is separable, and there is no need to exploit the quantum probabilistic interpretation. Therefore, through the concept of nonseparability, the EFNN can capture both separable and nonseparable bimodal interactions as a generalization of existing probabilistic modality fusion approaches. This attribute is the core reason that EFNN has achieved remarkable performance improvements over a range of traditional and quantum-inspired baselines.

Another work taking inspiration from quantum entanglement is QMF~\cite{Li2021}. The intramodal interaction and the interaction across different modalities (i.e., text, image and sound) are formulated with superposition and entanglement, respectively. Examples of intramodal interactions here are sentence structures and word dependencies for text or evolution of human facial expressions over time within a specific modality. Intermodal interaction refers to the interactions between different modalities on a higher level.
The interactions between different words in an individual modality are captured by quantum superposition on the feature level.
The intermodal interaction is akin to quantum entanglement. Sentiment is perceived via a complex process for understanding textual, visual and acoustic information, and this process cannot be regarded as a simple combination of unimodal clues.

Specifically, a multimodal Hilbert space is computed by the tensor product of a set of basis states, which are textual, visual and acoustic Hilbert space basis states. One advantage of this kind of construction is that it can be adapted to any number of modalities. A word in the multimodal Hilbert space is a pure state. It is associated with a textual, visual and acoustic feature vector. Since a state is represented in a complex-valued manner, the phase-amplitude form of complex numbers was adopted. For the text modality, the amplitudes are constructed from the pretrained word embeddings through a deep neural network. The amplitudes of visual and acoustic modalities are obtained via two other deep neural networks from the respective input feature vectors. For the phases of each modality, sentiment polarity is used to initialize arguments for text modality and trained end-to-end. To obtain the representation of a sentence, global weighting and local mixture are used. After the mixture process, a sentence is represented by several local context states, each of which is a measure to yield a probability vector. All the probability vectors of one sentence form a probability matrix, which is passed to row-wise maximum pooling and a neural network to obtain the final sentiment classification result.
The model was also tested on the CMU-MOSI and CMU-MOSEI datasets and compared with seven baseline models. Even though the proposed method did not outperform the then-SOTA model (TFN~\cite{Zadeh2017}), it achieved a comparable performance.

It is worth noting that this work introduced the concept of a reduced density matrix, which allows us to understand how the unimodal judgments compose the final sentiment judgment. Specifically, by means of a reduced density matrix, we can generate sentiment prediction results for any bimodal and unimodal features, which is hard to achieve by classical tensor network models. However, the theoretical contribution of this work is largely undermined by the fact that quantum entanglement has not been quantitatively studied. The tensor product also caps the dimensionality of each modality, setting a limit to model effectiveness.

\subsection{Addressing Interactions of Conversationality and Multimodality: Quantum-inspired Multimodal Conversational Sentiment Analysis}
\begin{shaded}
\textit{\textbf{Task Definition: Multimodal Conversational Sentiment Analysis}. The model input is a multimodal\footnotemark\ conversation $S$, which consists of $N$ utterances $\{u_j\}^N_{i=1}$ and the corresponding sentiment labels $\{y_j\}_{i=1}^N$. Each utterance $u_j$ has textual, visual and acoustic representations $u^t_j$, $u^v_j$ and $u^a_j$. The objective of the task is to establish a function, mapping each multimodal utterance $u_j$ to its corresponding sentiment label $y_j$. }
\end{shaded}
\footnotetext{Here, we take a model with three modalities, i.e., textual, visual and acoustic, as an example.}

The multimodal conversational sentiment analysis task faces challenges from both the aforementioned conversational sentiment analysis and multimodal sentiment analysis tasks.
It needs to capture both contextual and cross-modal interactions, making it more challenging. However, researchers have found the quantum-cognitive perspective helpful to address this challenge.

By leveraging various QP formalisms and the LSTM architecture, ~\cite{Zhang2020} proposed a QMN framework. As an extension of the QIN model ~\citep{Zhang2019a} introduced earlier, QMN adds multimodal information to the conversational sentiment analysis. First, QMN uses a density matrix-based CNN (DM-CNN) subnetwork to extract and represent multimodal features for all utterances in a video. The strong-weak influence model in the QIN was kept to measure influence between speakers. The resultant influence matrix is integrated into the output gate of each LSTM cell. Then, with textual and visual features as input, the QMN uses two separate LSTM networks to obtain their hidden states, which are then fed into the softmax function to obtain unimodal sentiment analysis results. Finally, a multimodal decision fusion method inspired by quantum interference is devised to derive final decisions based on unimodal results.

The model was evaluated on two widely used datasets, Multimodal EmotionLines Dataset (MELD) and IEMOCAP. For the MELD dataset, on both the 3-class sentiment analysis task and the 7-class emotion recognition task, the QMN obtained the best result. The 5-class sentiment analysis task on the IEMOCAP dataset also achieved the best performance with an approximately 3.6\% boost over the hierarchical contextual h-LSTM network~\cite{Poria2017}.
However, the limitation of the work lies in the fact that more concepts from QT can be exploited to model interactions, such as quantum composite systems, quantum entanglement and quantum interference, to better use the modeling power of QT.

\cite{Li2021b} proposed a quantum-like framework to handle the multimodal conversational emotion recognition problem. 
In quantum physics experiments, particles are in a mixture of multiple independent pure states until they are measured, and the measurement will cause them to collapse into a single pure measurement state. 
Likewise, the speaker is initially in a mixture of multiple independent emotions. Treating the conversation context as a measurement can cause the emotional state to collapse to a pure state. Furthermore, the evolution of quantum states over time can be analogized to the evolution of a speaker's emotion state during a conversation. 
These analogies inspired the design of a quantum measurement-inspired model, namely, QMNN, for multimodal conversational emotion recognition. Multimodal fusion is achieved by a quantum mixture system, where features from a single modality are treated as pure states. Utterances are treated as a mixture of pure states. A quantum-like recurrent neural network is used to capture the evolution of the conversation context. Then, a complex-valued neural network is constructed to implement the measurement process. A dedicated optimizer is used to update complex-valued unitary matrices in the evolution process, allowing the entire model to be trained in an end-to-end style.

QMNN was also evaluated on the MELD and IEMOCAP datasets but with different classification tasks. On the 6-class emotion recognition task, the SOTA DialogueRNN model \cite{Ghosal2019} obtained the best result, but on the MELD for the 7-class emotion recognition task, QMNN outperformed all the baseline models.

Among all the above surveyed models, QMNN is the only model that deals with the fusion of three modalities. Using three modal vectors provides rich sentiment and semantic information and the utterance order and the speaker information are encoded in the phases of the complex-valued representation. As such, the expressive ability of quantum complex-valued representation can be better exerted. 
Additionally, different from the use of quantum evolution in the EFNN model ~\citep{Gkoumas2021a} discussed in the previous subsection, a model called the quasirecurrent neural network (QRNN) was proposed in \cite{Li2021b} for the quantum evolution of states. Compared with the traditional RNN, QRNN can better render the uncertainties in the conversational context with its hidden units. In addition, it has shown strong potential in memorizing contextual information. However, as QRNN deals with density matrices, it imposes a higher computational cost than the use of vectors. Therefore, only one QRNN layer was actually used, limiting the model effectiveness. In the future, a more efficient method of deploying QRNN should be investigated so that more layers can be applied to obtain better results.

\subsection{Addressing the Interrelatedness of Sentiment Aspects: Quantum-inspired Multimodal Conversational Multitask Sentiment Analysis}

 Since human sentiments are of different and usually interrelated types (such as sentiment, emotion and sarcasm), multimodal and conversational sentiment analysis can be naturally viewed as a multitask problem, posing the challenge of multitask decision-making. The main motivation behind multitask learning is that incorporating information from other related tasks may improve the target task. A common practice is to share parameters between different tasks to enhance the generalization ability of the target task. It is usually done by weighting the training information between the target task and auxiliary tasks. In quantum-inspired models, the formalism of incompatibility has been exploited to address this problem.
\begin{shaded}
\textit{\textbf{Task Definition: Multitask Sentiment Analysis}. The $j^{th}$ sample in the dataset is represented as $\{X_j=(C_{i=1}^m,T_j), Y_j=(task_1, task_2))\}$, where $C_j$, $T_j$ and $Y_j$ denote the context utterance, the target utterance and its sentiment labels, respectively. The last utterance in a conversation is regarded as the target utterance, and its preceding $m$ utterances are considered context. $task_1$ and $task_2$ correspond to the sentiment labels of the target utterance for two different tasks (e.g., emotion recognition and sarcasm recognition). Each utterance in the conversation $X_j$ is represented by textual, visual and acoustic features. The objective is to find a function to map from all the input utterances of a conversation to the target utterance's sentiment labels for $task_1$ and $task_2$.}
\end{shaded}

As an extension of the work~\cite{Zhang2021cfn}, \cite{Liu2021} proposed a multitask learning framework for joint multimodal sentiment and sarcasm classification. The motivation was that sentiment and sarcasm are closely related, and understanding one will benefit the other. Thus, jointly analyzing sarcasm and sentiment would improve the performance of both tasks. The authors proposed a quantum probability theory driven multitask (QPM) learning framework consisting of a complex-valued embedding layer, a quantum-like fusion network, a quantum measurement layer and a dense layer. Conversational interactions and intermodality interactions are captured by the quantum composite system and a quantum interference mechanism. The work further explored the intertask correlations through quantum incompatible measurements. It is important to stress that the paper also provided a set of theoretical justifications on why QPM is suitable for joint detection of sarcasm and sentiment.

As the only multitask framework thus far for quantum-inspired sentiment analysis, there was a lack of exactly fit-for-purpose datasets. In this work, the MUStARD-improved and MEMOTION datasets were chosen for empirical evaluation. Note that MUStARD-improved contains contextual information, but MEMOTION does not. 
QPM achieved the best performance, with improvements of 1\% and 2\% over the state-of-the-art (A-MTL \cite{Chauhan2020}) on each dataset.
 Extended experiments not only demonstrate the superior performance of the model over a range of baselines but also show the existence of incompatibility between different tasks.
A metric based on quantum relative entropy was then introduced to measure the degree of incompatibility between multiple tasks. This was an initial step toward investigating incompatible measurements in the multitask learning setting, creating room for further research in this field. 

\begin{table*}[th]
\begin{center}
\caption{Experimental results of the surveyed quantum-cognitively inspired sentiment analysis models. The table is categorized by the task that each model is designed for. Note that even though some models are categorized under one task, they may have used different datasets in different ways. }
\label{table:result}
\scalebox{0.78}{
\begin{tabular}{l|l|l|c|c|c|c}
\toprule
\textbf{Dataset}            & \textbf{Category}            & \textbf{Model}               & \textbf{Accuracy} & \textbf{Acc-improvement} & \textbf{F1} & \textbf{F1-improvement} \\ 
\midrule
\multirow{7}{*}{MELD}       & \multirow{3}{*}{sentiment-3} & QIN                          & 0.679             & 0.027                    & 0.662       & 0.024                   \\ \cline{3-7} 
                            &                              & DMATT-BiGRU                  & 0.525             & 0.027                    & -           & -                       \\ \cline{3-7} 
                            &                              & QMN                          & 0.756             & 0.049                    & 0.729       & 0.036                   \\ \cline{2-7} 
                            & \multirow{4}{*}{emotion-7}   & QIN                          & 0.619             & 0.011                    & 0.578       & 0.015                   \\ \cline{3-7} 
                            &                              & DMATT-BiGRU                  & 0.441             & 0.011                    & -           & -                       \\ \cline{3-7} 
                            &                              & QMN                          & 0.693             & 0.029                    & 0.627       & 0.012                   \\ \cline{3-7} 
                            &                              & QMNN                         & 0.6081            & 0.0016                   & 0.58        & 0.001                   \\ \hline
\multirow{3}{*}{IEMOCAP}    & emotion-9                    & QIN                          & 0.376             & 0.025                    & 0.343       & 0.008                   \\ \cline{2-7} 
                            & sentiment-5                  & QMN                          & 0.648             & 0.023                    & 0.623       & 0.021                   \\ \cline{2-7} 
                            & emotion-6                    & QMNN                         & 0.6084            & -0.0158                  & 0.5988      & -0.006                  \\ \hline
\multirow{5}{*}{CMU-MOSI}   & \multirow{2}{*}{sentiment-2} & Decision Fusion               & 0.846             & 0.062                    & 0.845       & 0.068                   \\ \cline{3-7} 
                            &                              & EFNN                         & 0.809             & 0.027                    & 0.808       & 0.029                   \\\cline{3-7} 
                            &                              & QMF                         & 0.8069             & -0.0049                    & -       & -                   \\ \cline{2-7} 
                            &\multirow{2}{*}{sentiment-7}   & EFNN                         & 0.359             & 0.028                    & -           & -                       \\ \cline{3-7} 
                            &                              & QMF                  & 0.4788          & -0.0105                    & -           & -                       \\ \hline
\multirow{5}{*}{CMU-MOSEI}  & \multirow{2}{*}{sentiment-2} & Decision Fusion               & 0.849             & 0.027                    & 0.911       & 0.03                    \\ \cline{3-7} 
                            &                              & EFNN                         & 0.828             & 0.026                    & 0.826       & 0.029                   \\\cline{3-7} 
                            &                              & QMF                         & 0.7974             &  -0.0028                   & -       & -                   \\ \cline{2-7} 
                            &\multirow{2}{*}{sentiment-7}   & EFNN                         & 0.502             & 0.036                    & -           & -                       \\ \cline{3-7} 
                            &                              & QMF                  & 0.3353             &  -0.0335                   & -           & -                 \\ \hline
\multirow{3}{*}{MUStARD}    & \multirow{2}{*}{sarcasm-2}   & CFN                          & 0.754             & 0.059                    & 0.754       & 0.061                   \\ \cline{3-7} 
                            &                              & QPM                          & 0.775             & -                        & 0.7753      & 0.013                   \\ \cline{2-7} 
                            & sentiment-3                  & QPM                          & 0.6605            & -                        & 0.6611      & -                       \\ \hline

\multirow{2}{*}{IMDB}    & sentiment-2                  & DMATT-BiGRU                  & 0.925             & 0.019                    & -           & -                       \\ \cline{2-7} 
                            & sentiment-2                  & QI-CNN                          & 0.7523             & 0.0084                    & 0.7405      & 0.0004                 \\ \hline
\multirow{3}{*}{IEMOCAP}    & emotion-9                    & QIN                          & 0.376             & 0.025                    & 0.343       & 0.008                   \\ \cline{2-7} 
                            & sentiment-5                  & QMN                          & 0.648             & 0.023                    & 0.623       & 0.021                   \\ \cline{2-7} 
                            & emotion-6                    & QMNN                         & 0.6084            & -0.0158                  & 0.5988      & -0.006                  \\ \hline

\multirow{2}{*}{ScenarioSA} & ScenarioSA(A)                & \multirow{2}{*}{DMATT-Bigru} & 0.667             & 0.057                    & -           & -                       \\ \cline{2-2} \cline{4-7} 
                            & ScenarioSA(B)                &                              & 0.738             & 0.023                    & -           & -                       \\ \hline
Reddit-Track                & sarcasm-2                    & CFN                          & 0.68              & -0.028                   & 0.68        & -0.028                  \\ \hline
OMD                & sentiment-2                    & QSR                          & 0.7775              & -0.0031                 & 0.7759        & -0.0025                  \\ \hline
Sentiment140                & sentiment-2                    & QSR                          & 0.7185              & 0.0177                   & 
0.7083        & 0.009                  \\ \hline
\multirow{2}{*}{MEMOTION}   & sarcasm-2                    & \multirow{2}{*}{QPM}         & 0.6145            & -                        & 0.6139      & 0.021                   \\ \cline{2-2} \cline{4-7} 
                            & sentiment-3                  &                              & 0.4267            & -                        & 0.427       & -                       \\ \hline
Flickr                      & sentiment-2                  & \multirow{2}{*}{QMSA}        & 0.9312            & 0.0208                   & 0.9301      & 0.0132                  \\ \cline{1-2} \cline{4-7} 
Getty Image                 & sentiment-3                  &                              & 0.8824            & 0.0365                   & 0.8969      & 0.0494                  \\ \hline
WeiBo                 & sentiment-2                  &  QI-CNN                            & 0.7147            & 0.0062                   & 0.7145      & 0.0044                  \\ 
\bottomrule
\end{tabular}}
\end{center}
\end{table*}

\section{Summary}

In this paper, we provided a comprehensive survey on the background, motivation, current status, main results and findings of the recently developed quantum-cognitively inspired sentiment analysis models. The principled superiority of QT in modeling human cognition and decision-making processes is the main driving force of the recent development of this research direction. Researchers have identified the potential fit of QT for addressing the main challenges of sentiment analysis, including \textit{ambiguity in sentiment judgment}, \textit{conversationality}, \textit{multimodality} and \textit{interrelatedness of different sentiment aspects}, and established conceptual mappings between QT and sentiment analysis. Based on the mappings, researchers have built computational models for implementing quantum-cognitive frameworks on sentiment analysis tasks. The computational models are mainly neural networks with complex-valued and numerically constrained layers to ensure quantumness. Table~\ref{table:result} summarizes the experimental results of the surveyed models as reported in the original papers. These quantum-cognitively inspired sentiment analysis models have achieved encouraging results in comparison with the state-of-the-art classical models while simultaneously demonstrating better interpretability over traditional DL models. \\

\noindent \textbf{Challenges Tackled.} The existing works have targeted the core challenges for sentiment analysis. In particular, the pure and mixed state representations compactly encode the semantics and sentiment in words and sentences, handling inherent ambiguity in sentiment judgment; the challenge of multimodal fusion is handled by interference or entanglement; the evolving conversational context is encoded in a sentiment analysis framework by the notions of quantum measurement and quantum evolution; and the interrelatedness between different sentiment aspects has been appropriately handled by quantum incompatibility. For each task, researchers have managed to convert the theoretical frameworks to real applicable models and devise approaches to make them trainable on large-scale and real-life data. \\

\noindent \textbf{Advantages.} The surveyed works have empirically supported the claims that QT can well formulate human cognition and decision-making processes. The experimental results suggest that quantum-cognitive models have achieved a good trade-off between effectiveness and theoretical soundness: most models were designed to ensure theoretical soundness in the first place, while some measures taken to execute them led to minor performance drops as a side effect, although some models have achieved significant performance improvements. In contrast, traditional sentiment analysis models do not take a predetermined human cognitive viewpoint and hence are hard to interpret with human language. The surveyed quantum-cognitive models act according to human cognitive views of sentiment and still achieve decent performance in real tasks. As demonstrated in many works, good explanations can be produced by post hoc analysis of these quantum models.

In addition, the benefits of complex values have been recognized and exploited in semantic-aware representation learning. Commonly, quantum-inspired models first leverage the Hilbert space representation to represent basic linguistic units as quantum states. The basic units are mostly words but sometimes can also be sememes or sentiment polarity. Phase-amplitude decomposition has provided a natural way to place semantics and sentiment in a coupled manner. Most surveyed papers have demonstrated that complex-valued models outperform their real counterparts with doubled dimensionalities, which confirms the effect of phase-amplitude encoding in practice.\\

\section{Limitations and Future Directions}

We have shown that quantum-cognitively inspired sentiment analysis models have achieved relatively good results and demonstrated strong capability in tackling the ambiguity, contextuality, and multimodality issues of sentiment analysis tasks. They have also been shown to offer improved interpretability. 

Despite the major theoretical breakthroughs and encouraging experimental results discussed in this paper, we believe that these models still have some limitations, generally in the following aspects: a) the fundamental methods, b) limited scenarios, and c) efficiency issues. These limitations are discussed in more detail below, highlighting corresponding future research directions to address them. We hope this will lay a sensible roadmap for this promising research area to move forward and make further breakthroughs. 

\begin{itemize}

    \item \textbf{Fundamental investigation of representation learning in complex-valued domains}
Complex-valued representation is an indispensable foundational concept in the mathematical formalism of QT. Existing quantum-inspired models use wave functions and polar coordinate equations to represent complex values. The amplitude part is approximated by pretrained word embedding vectors, while the imaginary part is often randomly initialized. Moreover, the DL toolkits used have not supported complex-valued representation until recently. Therefore, how to effectively implement and make use of the complex-valued representation is a direction worth exploring.

\item \textbf{Quantitative exploration of interference and entanglement}
The post hoc interpretability of quantum-inspired models has been studied in some of the surveyed papers through a correspondence of the core model components with their QT explanations or via case studies. Apart from post hoc interpretability, another aspect of interpretability (sometimes called `transparency') that targets uncovering the actual working mechanisms \cite{Lipton2016} is underinvestigated. Although certain preliminary explorations of the core concepts exist in QT (e.g., quantum interference, incompatibility, and entanglement), few of the existing works have quantitatively measured their effects. For example, the formalism of quantum relative entropy has been applied to give a quantitative analysis on the issue of incompatibility. More efforts need to be devoted to assessing the effectiveness of different components in quantum-cognitively inspired models. 
We expect more systematic and principled methods from QT to be explored to quantify these concepts, which would not only benefit interpretability but also help better analyze the performance and effectiveness of the models.

\item \textbf{Better compatibility with the existing paradigm}
The surveyed quantum-inspired models usually tailor the quantum formalization on the basis of the end-to-end neural network paradigm. As a consequence, there naturally exists some mismatch between quantum formalization and the end-to-end paradigm. For example, from a quantum state point of view, any object is represented as a unit-length state vector. However, textual features, which are universally discovered as the main information carrier of sentiment among these multiple-modality features, are represented as static word vectors that are usually not of unit length and, therefore, have to be converted to unit vectors to be compatible with a quantum state view.
The typical approach of this aim is to apply L2 normalization of a vector as the (amplitude of) input state. However, the L2 normalization of the pretrained word vectors might damage the semantic information stored in the vector, as the strategy is not inherently supported in pretraining the vectors. To improve the performance, one may investigate alternative approaches to transforming the pretrained word vectors to normalized input states or directly pretraining word vectors in a quantum-cognitive manner.

\item \textbf{More delicate user studies}
The surveyed works experimentally demonstrate the effectiveness of quantum-cognitively inspired models. Nonetheless, we believe that examining the quantum-like effects of sentimental analysis implicitly captured in these models is also essential. Such examinations are not limited to designing system experiments (typically run by computers). As a powerful tool in cognition and psychology, user studies might help in this regard. As \cite{Bruza2015,Wang2016} did, we expect more delicate user studies that borrow some protocols from cognition and psychology to be carried out to examine these quantum-like effects. 

\item \textbf{Solving efficiency issues}
The operations in the current quantum-cognitively inspired models are largely based on complex-valued density matrices, which are slightly different from typical neural networks and might be unfriendly in current hardware or computing libraries. For example, density matrices, with a space complexity of $\mathcal{O} (D^2)$, are commonly used as the basic components for textual (or multimodal) representations. Further operations based on density matrices are also quadratic in the feature dimensionality $D$, which is computationally expensive. We expect some efforts in hardware optimization or algorithmic progress to relieve the efficiency issues.

\item \textbf{Exploring richer scenarios and building benchmarks} To investigate quantum-cognitively inspired modeling in sentiment analysis, one could explore richer scenarios. These scenarios are twofold: data multimodality and multiple aspects of sentiment. First, with the development of the internet, mobile devices and social platforms, data of various types continue to grow in scale. These data are not limited to textual, visual, or acoustic data but also include other modalities such as fingerprints and gestures. To address such richer multimodality, more efficient fusion models are expected. 
Second, various aspects of sentiment can also be explored. We have already studied sentiment analysis, sarcasm detection and emotion recognition. Other aspects, such as hate speech detection, desire recognition and metaphor analysis, are expected to be explored. 
Furthermore, standardized benchmarks are basically the first driving force of modern ML, e.g., ImageNet \cite{Krizhevsky2012} and General Language Understanding Evaluation (GLUE) \cite{Wang2018a}. More high-quality multimodal datasets are needed, which involve not only multiple modalities but also multiple aspects of sentiment during annotation.  

\item \textbf{Possible adaptations on quantum computers}
As an interdisciplinary research field, one natural question is how quantum-cognitively inspired models for sentiment analysis could benefit from quantum computers. These surveyed models are currently run on classical computers using quantum formalization. 
Classical computers process classical bits, while quantum models are based on superposed states (qubits). It would be interesting to see if quantum computers could bring something new to quantum-inspired models. Executing NLP models on quantum computers is not far as researchers have already built and deployed models for toy NLP tasks on actual quantum hardware \cite{Coecke2020,Lorenz2021}. This encourages us to imagine a future when quantum-inspired models would be executed in quantum computers.

\end{itemize}
\normalem

\section*{Acknowledgments}
This research was supported in part by the Natural Science Foundation of Beijing (grant number:  4222036). We would like to thank the anonymous reviewers for their valuable comments and Dr. Sagar Uprety for his kind help in proofreading the manuscript. 

\bibliographystyle{ACM-Reference-Format}
\bibliography{Untitled}

\end{document}